\documentclass[sigconf]{acmart}

\AtBeginDocument{%
  \providecommand\BibTeX{{%
    \normalfont B\kern-0.5em{\scshape i\kern-0.25em b}\kern-0.8em\TeX}}}

\copyrightyear{2021}
\acmYear{2021}
\setcopyright{acmcopyright}\acmConference[KDD '21]{Proceedings of the 27th ACM SIGKDD Conference on Knowledge Discovery and Data Mining}{August 14--18, 2021}{Virtual Event, Singapore}
\acmBooktitle{Proceedings of the 27th ACM SIGKDD Conference on Knowledge Discovery and Data Mining (KDD '21), August 14--18, 2021, Virtual Event, Singapore} \acmPrice{15.00}
\acmDOI{10.1145/3447548.3467350}
\acmISBN{978-1-4503-8332-5/21/08}
\usepackage{bm}
\usepackage{booktabs}
\usepackage{makecell}
\usepackage{multirow}
\usepackage{paralist}
\usepackage{enumitem}
\setlist[itemize]{leftmargin=3.5mm}
\setlist[enumerate]{leftmargin=5mm}

\definecolor{comment}{RGB}{64, 130, 39}
\definecolor{purple2}{RGB}{136, 72, 246}
\definecolor{linkcolor}{RGB}{0,0,238}

\newcommand{\yx}[1]{\textit{{\color{red} YX: #1 }}}

\DeclareMathOperator*{\concat}{\scalebox{1}[1.5]{$\parallel$}}
\begin{document}
\fancyhead{}

\title{Are we really making much progress? Revisiting, benchmarking, and refining  heterogeneous graph neural networks}

\author[Q. Lv*, M. Ding*, Q. Liu, Y. Chen, W. Feng, S. He, C. Zhou, J. Jiang, Y. Dong, J. Tang]{Qingsong Lv$^{*\dagger}$, Ming Ding$^{*\dagger}$, Qiang Liu$^{\clubsuit}$, Yuxiang Chen$^{\dagger}$, Wenzheng Feng$^{\dagger}$, Siming He$^{\diamondsuit}$, }
\author{Chang Zhou$^{\ddagger}$, Jianguo Jiang$^{\clubsuit}$, Yuxiao Dong$^{\mathparagraph}$, Jie Tang$^{\dagger \S}$}
\affiliation{
    $^\dagger$ Tsinghua University \country{}
    $^{\clubsuit}$ Chinese Academy of Sciences\country{}
    $^\ddagger$ Alibaba Group     \country{}
    $^{\diamondsuit}$ University of Pennsylvania\country{} $^\mathparagraph$ Microsoft 
}
\email{{lqs19, dm18, chenyuxi18, fwz17}@mails.tsinghua.edu.cn, {liuqiang, jiangjianguo}@iie.ac.cn}
\email{ericzhou.zc@alibaba-inc.com, siminghe@seas.upenn.edu, ericdongyx@gmail.com, jietang@tsinghua.edu.cn}

\renewcommand{\authors}{Qingsong Lv, Ming Ding, Qiang Liu, Yuxiang Chen, Wenzheng Feng, Siming He, Chang Zhou, Jianguo Jiang, Yuxiao Dong, Jie Tang}

\newcommand{\upsym}{\textcolor{comment}{$\bm{\uparrow}$}}
\newcommand{\dwsym}{\textcolor{red}{$\bm{\downarrow}$}}

\renewcommand{\shortauthors}{Lv and Ding, et al.}
\newcommand\blueit[1]{\textcolor{blue}{#1}}

\newcommand{\model}{Simple-HGN}
\newcommand{\smodel}{\model\space}
\def\*#1{\bm{#1}}
\def\&#1{\mathcal{#1}}
\newcommand{\vpara}[1]{\vspace{0.05in}\noindent\textbf{#1 }}
\newcommand{\hide}[1]{} 

\begin{abstract}
\renewcommand{\thefootnote}{\fnsymbol{footnote}}
\footnotetext[1]{Equal contribution.}
\footnotetext[4]{Corresponding Author.}
\renewcommand{\thefootnote}{\arabic{footnote}}

    Heterogeneous graph neural networks (HGNNs) have been blossoming in recent years, but the unique data processing and evaluation setups used by each work obstruct a full understanding of their advancements. 
    In this work, we present a systematical reproduction of 12 recent HGNNs by using their official codes, datasets, settings, and hyperparameters, revealing surprising findings about the progress of HGNNs.   
    We find that the simple homogeneous GNNs, e.g., GCN and GAT, are largely underestimated due to improper settings. 
    GAT with proper inputs can generally match or outperform all existing HGNNs across various scenarios.  
    To facilitate robust and reproducible HGNN research, we construct the Heterogeneous Graph Benchmark (HGB)\footnote{All codes and data are available at \textcolor{linkcolor}{\url{https://github.com/THUDM/HGB}}, and the HGB leaderboard is at \textcolor{linkcolor}{\url{https://www.biendata.xyz/hgb}}.}, consisting of 11 diverse datasets with three tasks. 
    HGB standardizes the process of heterogeneous graph data splits, feature processing, and performance evaluation. 
    Finally, we introduce a simple but very strong baseline \model ---which significantly outperforms all previous models on HGB---to accelerate the advancement of HGNNs in the future.  
    
\end{abstract}

\hide{
\begin{abstract}
    Heterogeneous Graph Neural Networks (HGNNs) are blossoming in recent years, but the diverse datasets and baselines obstruct a fully comparison among them. 
    In this work, we carefully reproduce 12 HGNNs and systematically analyze their defects. We find that the homogeneous GNNs, e.g., GCN and GAT, are largely underestimated due to improper settings. GAT with proper inputs basically matches or outperforms nearly all existing HGNNs. Some HGNNs even have problems like data leakage or tuning on test set. 
    
    We then construct the Heterogeneous Graph Benchmark (HGB), consisting of 11 datasets to fully compare the HGNNs. 
    
    We also propose a simple but very strong baseline \model, which significantly outperforms all previous models on HGB, to accelerate the advancement of HGNNs in the future.  
\end{abstract}

}



\maketitle




\section{Introduction}


As graph neural networks (GNNs)~\cite{gcn, battaglia2018relational} have already occupied the centre stage of graph mining research within recent years, the researchers begin to pay attention to their potential on heterogeneous graphs (a.k.a., Heterogeneous Information Networks)~\cite{Hu:KDD20GPT,dong2017metapath2vec,han,gtn,magnn,yang2020multisage}. Heterogeneous graphs consist of multiple types of nodes and edges with different side information, connecting the novel and effective graph-learning algorithms to the noisy and complex industrial scenarios, e.g., recommendation.

To tackle the challenge of heterogeneity,
various heterogeneous GNNs (HGNNs)~\cite{han,gtn,yang2020multisage} have been proposed to address the relevant tasks, including node classification, link prediction, and knowledge-aware recommendation. 
Take node classification for example, numerous HGNNs, such as HAN~\cite{han}, GTN~\cite{gtn}, RSHN~\cite{rshn}, HetGNN~\cite{hetgnn}, MAGNN~\cite{magnn}, HGT~\cite{hu2020heterogeneous}, and HetSANN~\cite{hong2020attention} were developed within the last two years. 

Despite various new models developed, our understanding of how they actually make progress has been thus far limited by the unique data processing and settings adopted by each of them.  
To fully picture the advancements in this field, we comprehensively reproduce the experiments of 12 most popular HGNN models by using the codes, datasets, experimental settings, hyperparameters released by their original papers. 
Surprisingly, we find that the results generated by these state-of-the-art HGNNs are not as exciting as promised (Cf. Table~\ref{tab:homo_comp}), that is:
\begin{enumerate}
    \item The performance of simple homogeneous GNNs, i.e., GCN~\cite{gcn} and GAT~\cite{velivckovic2017graph}, is largely underestimated. 
    Even vanilla GAT can outperform existing HGNNs in most cases with proper inputs.
    \item Performances of some previous works are mistakenly reported due to inappropriate settings or data leakage.
\end{enumerate}
Our further investigation also suggests:
\begin{enumerate}
\setcounter{enumi}{2} 
    \item Meta-paths are not necessary in most heterogeneous datasets. 
    \item There is still considerable room for improvements in HGNNs.
\end{enumerate}
In our opinion, the above situation occurs largely because the individual data and experimental setup by each work obstructs a fair and consistent validation of different techniques, thus greatly hindering the advancements of HGNNs.

To facilitate robust and open HGNN developments, we build the \textsc{Heterogeneous Graph Benchmark} (HGB). 
HGB currently contains 11 heterogeneous graph datasets that vary in {heterogeneity} (the number of node and edge types),  tasks (node classification, link prediction, and knowledge-aware recommendation), and {domain} (e.g., academic graphs, user-item graphs, and knowledge graphs). 
HGB provides a unified interface for data loading, feature processing, and evaluation, offering a convenient and consistent way to compare HGNN models. 
Similar to OGB~\cite{hu2020open}, HGB also hosts a leaderboard (\textcolor{linkcolor}{\url{https://www.biendata.xyz/hgb}}) for publicizing reproducible state-of-the-art HGNNs. 


Finally, inspired by GAT's significance in Table~\ref{tab:homo_comp}, we take GAT as backbone to design an extremely simple HGNN model---\model. 
\smodel can be viewed as GAT enhanced by three existing techniques: 
(1) {learnable type embedding} to leverage type information, 
(2) {residual connections} to enhance modeling power, 
and 
(3) ${L_2}$ {normalization on the output embedddings}. 
In ablation studies, these techniques steadily improve the performance. 
Experimental results on HGB suggest that \smodel can consistently outperform previous HGNNs on three tasks across 11 datasets, making it to date the first HGNN model that is significantly better than the vanilla GAT. 


To sum up, this work makes the following contributions: 

\begin{itemize}
    \item We revisit HGNNs and identify issues blocking progress; 
    \item We benchmark HGNNs by HGB for robust developments; 
    \item We refine HGNNs by designing the \smodel model. 
\end{itemize}


\begin{table*}[]
\centering
\small
\caption{Reproduction of Heterogeneous GNNs with simple GCN and GAT as baselines---all reproduction experiments use official codes  and the same dataset, settings, hyperparameters as the original paper. 
The line with star (*) are results reported in the paper, and the lines without star are our reproduction. ``-'' means the results are not reported in the original paper. We mark the reproduction terms with $>$1 point gap compared to the reported results by \upsym \ and \dwsym. We also keep the standard variance terms above 1.}

\label{tab:homo_comp}
\setlength{\tabcolsep}{0.7mm}
\begin{tabular}{c|cc|ccc|ccc|cccc|cc}
\toprule
        & \multicolumn{2}{c|}{HAN~\cite{han}}  & \multicolumn{3}{c|}{GTN~\cite{gtn}}    & \multicolumn{3}{c|}{RSHN~\cite{rshn}} & \multicolumn{4}{c|}{HetGNN~\cite{hetgnn}}                                 & \multicolumn{2}{c}{MAGNN~\cite{magnn}}   \\ 
\midrule
Dataset & \multicolumn{2}{c|}{ACM}  & DBLP  & ACM   & IMDB       & AIFB    & MUTAG  & BGS    & \multicolumn{2}{c}{MC (10\%)}   & \multicolumn{2}{c|}{MC (30\%)}  & \multicolumn{2}{c}{DBLP}  \\ 
\midrule
Metric  & {\footnotesize Macro-F1}   & {\footnotesize Micro-F1}     &   {\footnotesize Macro-F1}  &    {\footnotesize Macro-F1}     &     {\footnotesize Macro-F1}        &     {\footnotesize Accuracy}    &    Accuracy    &   Accuracy     & {\footnotesize Macro-F1}    & {\footnotesize Micro-F1}         & {\footnotesize Macro-F1}   & {\footnotesize Micro-F1}           & {\footnotesize Macro-F1}      & {\footnotesize Micro-F1} \\ 
\midrule
model*  & 91.89   & 91.85   & 94.18       & 92.68  & 60.92        & 97.22   & 82.35  & 93.10  & 97.8     & 97.9   & 98.1                           & 98.2                           & 93.13                                    & 93.61                                    \\ 
GCN*    & 89.31    & 89.45    & 87.30    & 91.60          & 56.89                                    & -       & -      & -      & -                               & -                               & -                               & -                               & 88.00                                    & 88.51                                    \\ 
GAT*    & 90.55                          & 90.55                          & 93.71                                    & 92.33                                    & 58.14                        & 91.67   & 72.06  & 66.32  & 96.2                           & 96.3                           & 96.5                           & 96.5                           & 91.05                                    & 91.61                                    \\ 
\midrule
model   &   90.94                                      &  90.96                                       & 92.95\dwsym                          & 92.28                          & 57.53$\pm$2.22\dwsym                           & 97.22       & \textbf{82.35}      & 93.10      & 97.06            & 97.11     &  97.34         & 97.37                               & 92.81                           & 93.36                          \\ 
GCN     & \textbf{92.25}\upsym     & \textbf{92.29}\upsym       & 91.48\upsym                        & 92.28             & \textbf{59.11$\pm$1.73}\upsym & 97.22   & 79.41  & 96.55  & 91.88                           & 92.04    & 95.37 & 95.57                           & 88.31                        & 89.37                         \\ 
GAT     & 92.08\upsym & 92.15\upsym & \textbf{94.18} & \textbf{92.49} & 58.86$\pm$1.73                           & \textbf{100}\upsym     & 80.88\upsym  & \textbf{100}\upsym    & \textbf{98.25}\upsym & \textbf{98.30}\upsym & \textbf{98.42}\upsym & \textbf{98.50}\upsym & \textbf{94.40}\upsym & \textbf{94.78}\upsym \\ \bottomrule
\end{tabular}
\end{table*}

\section{Preliminaries}\label{sec:prelim}
\subsection{Heterogeneous Graph}

A heterogeneous graph~\cite{Sun:BOOK2012} can be defined as $G=\{V, E, \phi, \psi\}$, where $V$ is the set of nodes and $E$ is the set of edges. Each node $v$ has a type $\phi(v)$, and each edge $e$ has a type $\psi(e)$. The sets of possible node types and edge types are denoted by $T_v=\{\phi(v):\forall v\in V\}$ and $T_e=\{\psi(e):\forall e\in E\}$, respectively. When $|T_v| =  |T_e| = 1$, the graph degenerates into an ordinary \emph{homogeneous} graph. 

\subsection{Graph Neural Networks} 
GNNs aim to learn a representation vector $\*h_v^{(L)} \in \mathbb{R}^{d_L}$ for each node $v$ after L-layer transformations, based on the graph structure and the initial node feature $\*h^{(0)}_v \in \mathbb{R}^{d_0}$. The final representation can serve various downstream tasks, e.g., node classification, graph classification (after pooling), and link prediction. 

\textbf{Graph Convolution Network} (GCN)~\cite{gcn} is the pioneer of GNN models, where the $l^{th}$ layer is defined as
\begin{equation}
    \*H^{(l)} = \sigma(\hat{\*A}\*H^{(l-1)}\*W^{(l)}),
\end{equation}
where $\*H^{(l)}$ is the representation of all nodes after the $l^{th}$ layer. 
$\*W^{(l)}$ is a trainable weight matrix. 
$\sigma$ is the activation function, 
and $\hat{\*A}$ is the normalized adjacency matrix with self-connections. 

\textbf{Graph Attention Network} (GAT)~\cite{velivckovic2017graph} later replaces the average aggregation from neighbors, i.e., $\hat{\*A}\*H^{(l-1)}$, as a weighted one, where the weight $\alpha_{ij}$ for each edge $\langle i,j\rangle$ is from an attention mechanism as (layer mark $(l)$ is omitted for simplicity)
\begin{equation}
    \alpha_{ij} = \frac{\exp\left(\text{LeakyReLU}\left(\*a^T[\*W\*h_i\|\*W\*h_j]\right)\right)}{\sum_{k\in \&N_i} \exp\left(\text{LeakyReLU}\left(\*a^T[\*W\*h_i\|\*W\*h_k]\right)\right)},
\end{equation}
where $\*a$ and $\*W$ are learnable weights and $\&N_i$ represents the neighbors of node $i$. Multi-head attention technique~\cite{vaswani2017attention} is also used to improve the performance.

Many following works~\cite{abu2019mixhop, xu2018powerful, ding2018semi, feng2020graph} improve GCN and GAT furthermore, with focuses on homogeneous graphs. 
Actually, \textit{these homogeneous GNNs can also handle heterogeneous graphs by simply ignoring the node and edge types. }

\subsection{Meta-Paths in Heterogeneous Graphs}

Meta-paths~\cite{Sun:BOOK2012,sun2011pathsim} have been widely used for mining and learning with heterogeneous graphs. A meta-path is a path with a pre-defined (node or edge) types pattern, i.e., $\&P \triangleq n_1 \xrightarrow{r_1} n_2 \xrightarrow{r_2} \cdots \xrightarrow{r_l} n_{l+1}$, where $r_i\in T_e$ and $n_i\in T_v$. Researchers believe that these composite patterns imply different and useful semantics. For instance, ``author$\leftrightarrow$paper$\leftrightarrow$author'' meta-path defines the ``co-author'' relationship, and ``user$\xrightarrow{\text{buy}}$item$\xleftarrow{\text{buy}}$user$\xrightarrow{\text{buy}}$item'' indicates the first user may be a potential costumer of the last item. 

Given a meta-path $\&P$, we can re-connect the nodes in $G$ to get a \textbf{meta-path neighbor graph} $G_{\&P}$. Edge $u\rightarrow v$ exists in $G_{\&P}$ if and only if there is at least one path between $u$ and $v$ following the meta-path  $\&P$ in the original graph $G$.

\hide{

\section{Introduction}
As Graph Neural Nets (GNNs)~\cite{gcn, battaglia2018relational} have already occupied the centre stage of graph mining research within recent years, the researchers begin to pay attention to their potential on heterogeneous graphs (a.k.a. Heterogeneous Information Networks)~\cite{han,gtn,magnn,yang2020multisage}. Heterogeneous graphs consist of multiple types of nodes and edges with different side information, connecting the novel and effective graph-learning algorithms to the noisy and complex industrial scenarios, e.g., recommendation. Due to the diversity of data and tasks, diverse methods are used to adapt GNNs to heterogeneous graphs, including meta-path~\cite{magnn,han,gtn}, novel attention mechanisms~\cite{han, hu2020heterogeneous, hong2020attention}, multi-task learning~\cite{hong2020attention, wang2019kgat}, and so on.  

To figure out the most effective techniques, we firstly survey and reproduce most previous Heterogeneous Graph Neural Nets (HGNNs). We surprisingly find that \begin{enumerate}
    \item The performance of traditional homogeneous GNNs, e.g. GCN~\cite{gcn} and GAT~\cite{velivckovic2017graph}, are largely underestimated. Even vanilla GAT without ``type information'' can outperform most previous HGNNs if properly used, as showed in the reproduction results Table~\ref{tab:homo_comp}.
    \item Results in some previous works are falsely reported due to inappropriate settings or data leakage.
    \item Meta-paths are not necessary in most current datasets.
    \item There is still considerable room for improvement in HGNNs. 
\end{enumerate}

We think the main reason for this situation is that various datasets and settings caused difficulties for comparison, and then hindered the advancement of HGNNs. For this sake, we collect and construct a new benchmark, \textsc{Heterogeneous Graph Benchmark} (HGB). 
HGB consists of 11 datasets from 3 tasks, i.e. node classification, link prediction and knowledge-aware recommendation on heterogeneous graphs.  
The datasets vary in \emph{scale} (the number of nodes and edges), \emph{heterogeneity} (the number of node and edge types) and \emph{domain} (such as academic graphs, user-item graphs and knowledge graphs). HGB provides a unified interface for data loading and evaluation, offering a convenient way to compare HGNN models. Besides, HGB use a unified three-stage pipeline, disentangling features preprocessing and downstream modules with HGNN to fairly compare the modeling power of the HGNN itself. Similar to GLUE~\cite{wang2018glue} and OGB~\cite{hu2020open}, HGB also hosts a leaderboard\footnote{The website and codes for all experiments will be released 
upon the end of double-blind review.} for publicizing reproducible state-of-the-art HGNNs. 

To establish a strong baseline for HGNNs research in the future, we design an effective and extremely simple model, named \model. \smodel can be viewed as GAT with three add-ons: (1) \textbf{learnable type embedding} to leverage type information, (2) \textbf{residual connections} to enhance modeling power, and (3) $\mathbf{L_2}$ \textbf{normalization on the output embedddings}. In ablation studies, the latter two techniques steadily improve the performance. Overall experiments show that \smodel outperforms previous models by a large margin (2\%$\sim$ 10\%) on HGB. 

This paper is organized as follows. Section~\ref{sec:prelim} introduces preliminaries about heterogeneous graph, GNN and meta-path. Section ~\ref{sec:relwork} ``dissects'' previous works, analyzing each work in details based on our findings when reproducing them. Section~\ref{sec:hgb} and \ref{sec:model} describe HGB and \smodel respectively. Finally, we discuss and conclude this paper in Section~\ref{sec:discuss}.

\section{Preliminaries}\label{sec:prelim}
\subsection{Heterogeneous Graph}

A heterogeneous graph can be defined as $G=\{V, E, \phi, \psi\}$, where $V$ is the set of nodes and $E$ is the set of edges. Each node $v$ has a type $\phi(v)$, and each edge $e$ has a type $\psi(e)$. The sets of possible node types and edge types are denoted by $T_v=\{\phi(v):\forall v\in V\}$ and $T_e=\{\psi(e):\forall e\in E\}$, respectively. When $|T_v| =  |T_e| = 1$, the graph degenerates into an ordinary \emph{homogeneous} graph. 

\subsection{Graph Neural Networks} 
GNNs aim to learn a representation vector $\*h_v^{(L)} \in \mathbb{R}^{d_L}$ for each node $v$ after L-layer transformations, based on the graph structure and the initial node feature $\*h^{(0)}_v \in \mathbb{R}^{d_0}$. The final representation can serve various downstream tasks, e.g., node classification, graph classification (after pooling) and link prediction. 

\textbf{Graph Convolution Network} (GCN)~\cite{gcn} is the pioneer of GNN models, where the $l^{th}$ layer is defined as
\begin{equation}
    H^{(l)} = \sigma(\hat{A}H^{(l-1)}W^{(l)}),
\end{equation}
where $H^{(l)}$ is the representation of all nodes after the $l^{th}$ layer, and $W^{(l)}$ is a trainable weight matrix. $\sigma$ and $\hat{A}$ are activation function and the normalized adjacency matrix with self-connections. 

\textbf{Graph Attention Network} (GAT)~\cite{velivckovic2017graph} then replaces the average aggregation from neighbors, i.e. $\hat{A}H^{(l-1)}$, as a weighted one, where the weight $\alpha_{ij}$ for each edge $\langle i,j\rangle$ are from an attention mechanism as follows (layer mark $(l)$ is omitted for simplicity),
\begin{equation}
    \alpha_{ij} = \frac{\exp\left(\text{LeakyReLU}\left(\*a^T[W\*h_i\|W\*h_j]\right)\right)}{\sum_{k\in \&N_i} \exp\left(\text{LeakyReLU}\left(\*a^T[W\*h_i\|W\*h_k]\right)\right)},
\end{equation}
where $\*a$ and $W$ are learnable weights and $\&N_i$ represents the neighbors of node $i$. Multi-head attention technique~\cite{vaswani2017attention} is also used to improve the performance.

Many following works~\cite{abu2019mixhop, xu2018powerful, ding2018semi, feng2020graph} improve GAT furthermore, but most of them only focus on homogeneous graphs, which motivates the HGNNs. Actually, these homogeneous GNNs can also handle heterogeneous graphs by ignoring the node and edge types. 

\subsection{Meta-paths in Heterogeneous Graph}

Meta-path~\cite{sun2011pathsim} is currently the dominated approach for tasks on heterogeneous graphs. A meta-path is a path with a pre-defined (node or edge) types pattern, i.e. $\&P \triangleq n_1 \xrightarrow{r_1} n_2 \xrightarrow{r_2} \cdots \xrightarrow{r_l} n_{l+1}$, where $r_i\in T_e$ and $n_i\in T_v$. Researchers believe that these composite patterns imply different and useful semantics. For instance, ``author$\leftrightarrow$paper$\leftrightarrow$author'' meta-path defines the ``co-author'' relationship, and ``user$\xrightarrow{\text{buy}}$item$\xleftarrow{\text{buy}}$user$\xrightarrow{\text{buy}}$item'' indicates the first user may be a potential costumer of the last item. 

Given a meta-path $\&P$, we can re-connect the nodes in $G$ to get a \textbf{meta-path neighbor graph} $G_{\&P}$. Edge $u\rightarrow v$ exists in $G_{\&P}$ if and only if there is at least one path between $u$ and $v$ following the meta-path  $\&P$ in the original graph $G$. 

}

\section{Issues with Existing Heterogeneous GNNs}\label{sec:relwork}
We analyze popular heterogeneous GNNs (HGNNs) organized by the tasks that they aim to address. 
For each HGNN, the analysis will be emphasized on its defects found in the process of reproducing its result by \textbf{using its official code, the same datasets, settings, and hyperparameters as its original paper}, which is summarized in Table~\ref{tab:homo_comp}.

\subsection{Node Classification}

\subsubsection{HAN~\cite{han}.}
Heterogeneous graph attention network (HAN) is among the early attempts to tackle with heterogeneous graphs. 
Firstly, HAN needs multiple meta-paths selected by human experts. 
Then HAN uses a hierarchical attention mechanism to capture both \emph{node-level} and \emph{semantic-level} importance. 
For each meta-path, the node-level attention is achieved by a GAT on its corresponding meta-path neighbor graph. 
And the semantic-level attention, which gives the final representation, refers to a weighted average of the node-level results from all meta-path neighbor graphs. 

A defect of HAN is its unfair comparison between HAN and GAT. 
Since HAN can be seen as a weighted ensemble of GATs on \underline{many} meta-path neighbor graphs, a comparison with the vanilla GAT is essential to prove its effectiveness. However, the GCN and GAT baselines in this paper take only \underline{one} meta-path neighbor graph as input, losing a large part of information in the original graph, even though they report the result of the best meta-path neighbor graph. 

To make a fair comparison, we feed the original graph into GAT by ignoring the types and only keeping the features of the target-type nodes. We find that \textit{this simple homogeneous approach consistently outperforms HAN, suggesting that the homogeneous GNNs are largely underestimated} (See Table~\ref{tab:homo_comp} for details).


Most of the following works also follow  HAN's setting to compare with homogeneous GNNs, suffering from the \textit{``information missing in homogeneous baselines''} problem, which leads to a positive cognitive deviation on the performance progress of HGNNs.

\subsubsection{GTN~\cite{gtn}.}
Graph transformer network (GTN) is able to discover valuable meta-paths automatically, instead of depending on manual selection like HAN. The intuition is that a meta-path neighbor graph can be obtained by multiplying the adjacency matrices of several sub-graphs. Therefore, GTN uses a soft sub-graph selection and matrix multiplication step to generate meta-path neighbor graphs, and then encodes the graphs by GCNs.

The main drawback of GTN is that it consumes gigantic amount of time and memory. For example, it needs 120 GB memory and 12 hours to train a GTN on DBLP with only 18,000 nodes. In contrast, GCN and GAT only take 1 GB memory and 10 seconds of time.

Moreover, when we test the GTN and GAT five times using the official codes of GTN, we find from Table~\ref{tab:homo_comp} that \textit{their average scores are not significantly different, though GTN consumes $>400\times$ time and $120\times$ memory of GAT.}

\subsubsection{RSHN~\cite{rshn}}

Relation structure-aware heterogeneous graph neural network (RSHN) builds coarsened line graph to obtain edge features first, then uses a novel Message Passing Neural Network (MPNN)~\cite{gilmer2017neural} to propagate node and edge features.

The experiments in RSHN have serious problems according to the official code. First, it does not use validation set, and just \textit{tune hyperparameters on test set}. 
Second, it \textit{reports the accuracy at the epoch with best accuracy on test set} in the paper. 
As shown in Table~\ref{tab:homo_comp}, our well-tuned GAT can even reach 100\% accuracy under this improper setting on the AIFB and BGS datasets, which is far better than the 91.67\% and 66.32\% reported in their paper.

\subsubsection{HetGNN~\cite{hetgnn}}

Heterogeneous graph neural network (HetGNN) first uses random walks with restart to generate neighbors for nodes, and then leverages Bi-LSTM to aggregate node features for each type and among types.

HetGNN has the same \textit{``information missing in homogeneous baselines''} problem 
as HAN: when comparing it with GAT, a sampled graph instead of the original full graph is fed to GAT. As demonstrated in Table~\ref{tab:homo_comp}, \textit{GAT with correct inputs gets clearly better performance.}

\subsubsection{MAGNN~\cite{magnn}}

Meta-path aggregated graph neural network (MAGNN) is an enhanced HAN. The motivation is that when HAN deals with meta-path neighbor graphs, it only considers two endpoints of the meta-paths but ignores the intermediate nodes. MAGNN proposes several meta-path encoders to encode all the information along the path, instead of only the endpoints.

However, there are two problems in the experiments of MAGNN. First, MAGNN inherits the \textit{``information missing in homogeneous baselines''} problem from HAN, and also \textit{underperforms GAT with correct inputs}.


More seriously, MAGNN has a \textit{data leakage} problem in link prediction, because it uses batch normalization, and loads positive and negative links sequentially during both training and testing periods. 
In this way, samples in a minibatch are either all positive or all negative, and the mean and variance in batch normalization will provide extra information. 
If we shuffle the test set to make each minibatch contains both positive and negative samples randomly, the AUC of MAGNN \textit{drops dramatically from 98.91 to 71.49} on the Last.fm dataset.

\subsubsection{HGT~\cite{hu2020heterogeneous}}
Heterogeneous graph transformer (HGT) proposes a transformer-based model for handling large academic heterogeneous graphs with heterogeneous subgraph sampling. 
As HGT mainly focuses on handling web-scale graphs via graph sampling strategy~\cite{hamilton2017inductive,yang2020understanding}, the datasets used in its paper (> 10,000,000 nodes) are unaffordable for most HGNNs, unless adapting them by subgraph sampling. 
To eliminate the impact of subgraph sampling techniques on the performance, 
we apply HGT with its official code on the relatively small datasets that are not used in its paper, producing mixed results when compared to GAT (See Table~\ref{tab:NC}). 

\subsubsection{HetSANN~\cite{hong2020attention}}
Attention-based graph neural network for heterogeneous structural learning (HetSANN) uses a type-specific graph attention layer for the aggregation of local information, avoiding manually selecting meta-paths. HetSANN is reported to have promising performance in the paper. 

However, the datasets and preprocessing details are not released with the official codes, and responses from its authors are not received as of the submission of this work. 
Therefore, we directly apply HetSANN with standard hyperparameter-tuning, giving unpromising results on other datasets (See Table~\ref{tab:NC}). 

\subsection{Link Prediction}

\subsubsection{RGCN~\cite{schlichtkrull2018modeling}}
Relational graph convolutional network (RGCN) extends GCN to relational (multiple edge types) graphs. The convolution in RGCN can be interpreted as a weighted sum of ordinary graph convolution with different edge types. For each node $i$, the $l^{th}$ layer of convolution are defined as follows,
\begin{equation}
    \*h_i^{(l)} = \sigma\left(\sum_{r\in T_e}\sum_{j\in \mathcal{N}_i^r}\frac{1}{c_{i,r}}\*W_r^{(l)}\*h_j^{(l)}+\*W_0^{(l)}\*h_i^{(l-1)} \right),
\end{equation}
where $c_{i,r}$ is a normalization constant and $W_0, W_r$s are learnable parameters. 


\subsubsection{GATNE~\cite{cen2019representation}}
General attributed multiplex heterogeneous network embedding (GATNE) leverages the graph convolution operation to aggregate the embeddings from neighbors. It relies on Skip-gram to learn a general embedding, a specific embedding and an attribute embedding respectively, and finally fuses all of them. In fact, GATNE is more a network embedding algorithm than a GNN-style model. 

\subsection{Knowledge-Aware Recommendation}
Recommendation is a main application for Heterogeneous GNNs, but most related works~\cite{fan2019metapath,fan2019graph,niu2020dual,liu2020kred} only focus on their specific industrial data, resulting in non-open datasets and limited transferability of the models.  
Knowledge-aware recommendation is 
an emerging sub-field, 
aiming to improve recommendation by linking items with entities in an open knowledge graph. In this paper, we mainly survey and benchmark models on this topic. 

\subsubsection{KGCN~\cite{kgcn} and KGNN-LS~\cite{kgnnls}} KGCN enhances the item representation by performing aggregations among its corresponding entity neighborhood in a knowledge graph. KGNN-LS further poses a
label smoothness assumption, which posits that similar items in the knowledge graph are likely to have similar user preference. It adds a regularization term to help learn such a personalized weighted knowledge graph.  
\subsubsection{KGAT~\cite{wang2019kgat}}\label{sec:kgat} KGAT shares a generally similar idea with KGCN. The main difference lies in an auxiliary loss for knowledge graph reconstruction and the pretrained BPR-MF~\cite{rendle2009bpr} features as inputs. Although not detailed in its paper, an important contribution of KGAT is to introduce the pretrained features into this tasks, which greatly improves the performance. Based on this finding, we successfully simplify KGAT and obtain similar or even better performance 
 (See Table~\ref{tab:recom_benchmark}, denoted as KGAT$-$). 

\subsection{Summary}

In summary, the prime common issue of existing HGNNs is the lack of fair comparison with homogeneous GNNs and other works---to some extent---encourage the new models to equip themselves with novel yet 
redundant modules, instead of focusing more on progress in performance. 
Additionally, a non-negligible proportion of works have individual issues, e.g., 
data leakage~\cite{magnn}, 
tuning on test set~\cite{rshn}, 
and 
two-order-of-magnitude more memory and time consumption without effectiveness improvements~\cite{gtn}. 

In light of the significant discrepancy, we take the initiative to setup a heterogeneous graph benchmark (HGB) with these three tasks on diverse datasets for open, reproducible heterogeneous graph research (See \S \ref{sec:hgb}). 
Inspired by the promising advantages of the simple GAT over dedicated and relatively-complex heterogeneous GNN models, we present a simple heterogeneous GNN model with GAT as backbone, offering promising results on HGB (See \S \ref{sec:model}).

\hide{

In summary, the prime common issue of existing HGNNs is the lack of fair comparison with homogeneous GNNs and other works, which to some extent encourage the new models to equip themselves with novel but redundant modules, instead of focusing more on progress in performance. Besides, a nonnegligible proportion of works have individual issues, e.g., data leakage in MAGNN~\cite{magnn}, tuning on test set in RSHN~\cite{rshn} and huge memory and time consumption in GTN~\cite{gtn}. 
}

\hide{

\section{Heterogeneous GNN Dissection}\label{sec:relwork}
In this section, we introduce three typical tasks on heterogeneous graphs. Previous HGNNs, organized by their corresponding tasks, are analyzed in details. Besides the core ideas of each model, we will emphasize their defects found in the process of reproduction, which is summarized in Table~\ref{tab:homo_comp}.

\subsection{Node Classification}

\subsubsection{HAN~\cite{han}.}
Heterogeneous graph attention network (HAN) is the first graph neural network to tackle with heterogeneous graphs specifically. 
Firstly, HAN needs multiple meta-paths selected by human. 
Then HAN uses a hierarchical attention mechanism to capture both \emph{node-level} and \emph{semantic-level} importance. For each meta-path, the node-level attention is achieved by a GAT on its corresponding meta-path neighbor graph. And the semantic-level attention, which gives the final representation, refers to a weighted average of the node-level results from all meta-path neighbor graphs. 

A 
defect of the HAN work is the unfair comparison between HAN and GAT. Since HAN can be seen as a weighted ensemble of GATs on many meta-path neighbor graphs, a comparison with vanilla GAT is essential to prove its effectiveness. However, the GCN and GAT baselines in this paper take only one meta-path neighbor graph as input, a large part of information in the original graph is lost, even though they report the result of the best meta-path neighbor graph. To make a fair comparison, we feed the original graph into GAT by ignoring the types and only keeping the features of the target-type nodes. We find that this approach steadily outperforms HAN, showing that the homogeneous GNNs are largely underestimated (See Table~\ref{tab:homo_comp} for details).


Unfortunately, most of the following work also follows the ``best meta-path neighbor graph'' setting to compare with homogeneous GNNs, which leads to a positive cognitive deviation on the performance progress of HGNNs.

\subsubsection{GTN~\cite{gtn}.}
Graph transformer network (GTN) is able to discover valuable meta-paths automatically, instead of depending on manual selection like HAN. The intuition is that a meta-path neighbor graph can be obtained by multiplying the adjacency matrices of several sub-graphs. Therefore, GTN uses a soft sub-graph selection and matrix multiplication step to generate meta-path neighbor graphs, and then encodes the graphs by GCNs.

The main drawback of GTN is that it consumes gigantic amount of time and memory. For example, it needs 120 GB memory and 12 hours to train a GTN on DBLP with only about 18,000 nodes. In contrast, GCN and GAT only take 1 GB memory and 10 seconds of time.

Moreover, when we test the GTN and GAT 5 times using the official codes of GTN, we find their average scores are not significantly different (Tale~\ref{tab:homo_comp}), though GTN consumes $>400\times$ time and $120\times$ memory of GAT.

\subsubsection{RSHN~\cite{rshn}}

Relation structure-aware heterogeneous graph neural network (RSHN) builds coarsened line graph to obtain edge features first, then uses a novel Message Passing Neural Network (MPNN)~\cite{gilmer2017neural} to propagate node and edge features.

The experiments in RSHN have serious problems according to the official codes. First, they do not use validation set, and just tune hyperparameters on test set. Besides, they report the accuracy at the epoch with best accuracy on test set as performance in the paper. As shown in Table~\ref{tab:homo_comp}, our well-tuned GAT can even reach 100 percent accuracy under this improper setting on the AIFB and BGS datasets, which is far better than the 91.67/66.32 reported in their paper.

\subsubsection{HetGNN~\cite{hetgnn}}

Heterogeneous graph neural network (HetGNN) first uses random walk with restart to generate neighbors for nodes, then uses Bi-LSTM to aggregate node features for each type and among types.

HetGNN has the same ``Information missing in homogeneous baselines'' problem as HAN, because when comparing their method with GAT, they feed a sampled graph instead of the original whole graph to GAT. As demonstrated in Table~\ref{tab:homo_comp}, GAT with correct inputs gets better performance.

\subsubsection{MAGNN~\cite{magnn}}

Meta-path aggregated graph neural network (MAGNN) is an enhanced HAN. The motivation is that when HAN deals with meta-path neighbor graphs, it only considers two endpoints of the meta-paths but ignores the intermediate nodes. MAGNN proposes several meta-path encoders to encode all the information along the path, instead of only the endpoints.

However, there are three problems in the experiments of MAGNN. Firstly, MAGNN also follows HAN's setting to compare with GCN and GAT, and still underperforms the GAT with correct inputs.

Besides, MAGNN modifies the IMDB dataset in an unreasonable way. IMDB is a multi-label classification dataset, to convert it to a single-label classification dataset, they only preserve the first label for each node. Since 28\% of nodes has more than one label, which may lead to a non-negligible negative influence on the evaluation of modeling power of HGNNs.

Last but more seriously, MAGNN has a data leakage problem in link prediction tasks, because it uses batch normalization, and load positive links and negative links sequentially during both training and testing periods. In this way, samples in a minibatch are either all positive or all negative, and the mean and variance in batch normalization will provide extra information. If we shuffle the test set to make each minibatch contains both positive and negative samples, the AUC of MAGNN drops dramatically from 98.91 to 71.49 on the Last.fm dataset.

\subsubsection{HGT~\cite{hu2020heterogeneous}}
Heterogeneous graph transformer (HGT) proposed a pipeline for large academic heterogeneous graphs, including heterogeneous subgraph sampling, relative temporal encoding, meta relation-aware heterogeneous mutual attention, heterogeneous message passing and target-specific heterogeneous message aggregation.   
As HGT mainly focuses on handling web-scale graphs via graph sampling strategy~\cite{hamilton2017inductive,yang2020understanding}, the datasets in its paper (> 10,000,000 nodes) are unaffordable for most HGNNs, unless adapting them by subgraph sampling. HGB contains relatively small graphs and can compare the efficacy of HGT with other HGNNs based on their original implementation.  

\subsubsection{HetSANN~\cite{hong2020attention}}
Attention-based graph neural network for heterogeneous structural learning (HetSANN) uses type-specific graph attention layer for the aggregation of local information, avoiding manually selecting meta-paths. HetSANN is reported to have promising performance in the paper. 

However, the datasets and preprocessing details are not released with the official codes, and we have not been able to contact the authors. Therefore, we cannot repeat their experiments. Directly applying HetSANN to HGB with conventional hyperparameter-tuning gives a bad result (See Table~\ref{tab:NC}), and we guess there might be some hidden details of vital importance.

\subsection{Link Prediction}

\subsubsection{RGCN~\cite{schlichtkrull2018modeling}}
Relational Graph Convolutional Network (RGCN) extends GCN to relational (multiple edge types) graphs. The convolution in RGCN can be interpreted as a weighted sum of ordinary graph convolution with different edge types. For each node $v$, the $l$-th layer of convolution are defined as follows,
\begin{equation}
    h_v^{(l)} = \sigma\left(\sum_{r\in T_e}\sum_{u\in \mathcal{N}_v^r}\frac{1}{c_{v,r}}W_r^{(l)}h_u^{(l)}+W_0^{(l)}h_v^{(l-1)} \right),
\end{equation}
where $c_{v,r}$ is a normalization constant and $W_0, W_r$s are learnable parameters. 


\subsubsection{GATNE~\cite{cen2019representation}}
General Attributed Multiplex HeTerogeneous Network Embedding (GATNE) leverages the graph convolution operation to aggregate the embeddings from neighbors. It relies on Skip-gram to learn a general embedding, a specific embedding and an attribute embedding respectively, and finally fuses all of them. In fact, GATNE is more a network embedding algorithm than a GNN-style model. In view of its generality and effectiveness, we include it in the benchmark to represent the typical results of network embedding algorithms.
\subsection{Knowledge-aware Recommendation}
Recommendation is a main application for HGNNs, but most related works~\cite{fan2019metapath,fan2019graph,niu2020dual,liu2020kred} only focus on their specific industrial data, resulting in non-open datasets, and limited transferability of the model.  
Knowledge-aware recommendation is a popular and relatively open sub-field, aiming to improve recommendation by linking items with entities in an open knowledge graph. In this paper, we mainly survey and benchmark models on this topic. 

\subsubsection{KGCN~\cite{kgcn} and KGNN-LS~\cite{kgnnls}} KGCN enhances the item representation by performing aggregations among its corresponding entity neighborhood in a knowledge graph. KGNN-LS further poses a
label smoothness assumption, which posits that similar items in the knowledge graph are likely to have similar user preference. It adds a regularization term to help learn such a personalized weighted knowledge graph.  
\subsubsection{KGAT~\cite{wang2019kgat}}\label{sec:kgat} KGAT shares a generally similar idea with KGCN. The only main difference is an auxiliary loss for knowledge graph reconstruction. However, we find that the pretrained BPR-MF~\cite{rendle2009bpr} features as inputs used in its official codes are the dominant factor for good performance. KGAT can obtain similar or even better performance after removing many other modules such as the TransE loss, attention mechanism and relation type information. (See Table~\ref{tab:recom_benchmark}, denoted as KGAT$-$). Therefore, we think KGAT may have some redundant designs.







}
\section{Heterogeneous Graph Benchmark}\label{sec:hgb}
\subsection{Motivation and Overview}\label{sec:motivation}

\vpara{Issues with current datasets.}
Several types of datasets---academic networks (e.g., ACM, DBLP), information networks (e.g., IMDB, Reddit), and recommendation graphs (e.g., Amazon, MovieLens)---are the most frequently-used datasets, but the detailed task settings could be quite different in different papers. 
For instance, HAN~\cite{han} and GTN~\cite{gtn} discard the citation links in ACM, while others use the original version. 
Besides, different splits of the dataset also contribute to uncomparable results. 
Finally, the recent graph benchmark OGB~\cite{hu2020open} mostly focuses on benchmarking graph machine  learning methods on homogeneous graphs and is not dedicated to heterogeneous graphs.

\vpara{Issues with current pipelines.} To fulfill a task, components outsides HGNNs can also play critical roles. 
For example, MAGNN~\cite{magnn} finds that not all types of node features are useful, and a pre-selection based on validation set could be helpful (See \S~\ref{sec:input}).  RGCN~\cite{schlichtkrull2018modeling} uses DistMult~\cite{yang2014embedding} instead of dot product for training in link prediction. 
We need to control the other components in the pipeline for fair comparison.

\vpara{HGB.}
In view of these practical issues, we present the heterogeneous graph benchmark (HGB) for open, reproducible heterogeneous GNN research. 
We standardize the process of data splits, feature processing, and performance evaluation, by establishing the HGB pipeline ``feature preprocessing $\rightarrow$ HGNN encoder $\rightarrow$ downstream decoder''. 
For each model, HGB selects the best fit \emph{feature preprocessing} and \emph{downstream decoder} based on its performance on validation set.  

\subsection{Dataset Construction}
HGB collects 11 widely-recognized \emph{medium-scale} datasets with \emph{predefined meta-paths} from previous works, making it available to all kinds of HGNNs. The statistics are summarized in Table~\ref{tab:statistics}.

\subsubsection{Node Classification}
Node Classification follows a transductive setting, where all edges are available during training and node labels are split according to 24\% for training, 6\% for validation and 70\% for test in each dataset.  
\begin{itemize}
    \item \textbf{DBLP}\footnote{\url{http://web.cs.ucla.edu/~yzsun/data/}} is a bibliography website of computer science. We use a commonly used subset in 4 areas with nodes representing authors, papers, terms and venues.
    \item \textbf{IMDB}\footnote{\url{https://www.kaggle.com/karrrimba/movie-metadatacsv}} is a website about movies and related information. A subset from Action, Comedy, Drama, Romance and Thriller classes is used.
    \item \textbf{ACM} is also a citation network. We use the subset hosted in HAN~\cite{han}, but preserve all edges including paper citations and references.
    \item \textbf{Freebase}~\cite{bollacker2008freebase} is a huge knowledge graph. We sample a subgraph of 8 genres of entities with about 1,000,000 edges following the procedure of a previous survey~\cite{yang2020heterogeneous}.
\end{itemize}

\subsubsection{Link Prediction} Link prediction is formulated as a binary classification problem in HGB. The edges are split according to 81\% for training, 9\% for validation and 10\% for test. Then the graph is reconstructed only by edges in the training set. For negative node pairs in testing, we firstly tried uniform sampling and found that most models could easily make a nearly perfect prediction (See Appendix~\ref{app:randneg}). Finally, we sample 2-hop neighbors for negative node pairs, of which are 1:1 ratio to the positive pairs in the test set.

\begin{itemize}
    \item \textbf{Amazon} is an online purchasing platform. We use the subset preprocessed by GATNE~\cite{cen2019representation}, containing electronics category products with co-viewing and co-purchasing links between them.
    \item \textbf{LastFM} is an online music website. We use the subset released by HetRec 2011~\cite{cantador2011second}, and preprocess the dataset by filtering out the users and tags with only one link.
    \item \textbf{PubMed}\footnote{https://pubmed.ncbi.nlm.nih.gov} is a biomedical literature library. We use the subset constructed by HNE~\cite{yang2020heterogeneous}.
\end{itemize}

\subsubsection{Knowledge-aware recommendation} We randomly split 20\% of user-item interactions as test set for each user, and for the left 80\% interactions as training set.

\begin{itemize}
    \item \textbf{Amazon-book} is a subset of Amazon-review\footnote{http://jmcauley.ucsd.edu/data/amazon/} related to books.
    \item \textbf{LastFM} is a subset extracted from last.fm with timestamp from January, 2015 to June, 2015.
    \item \textbf{Yelp-2018}\footnote{https://www.yelp.com/dataset} is a dataset adapted from 2018 edition of the Yelp challenge. Local businesses like restaurants and bars are seen as items.
    \item \textbf{Movielens} is a subset of Movielens-20M\footnote{https://grouplens.org/datasets/movielens/}, which is a widely used dataset for recommendation.
\end{itemize}

To assure the quality of dataset, we use 10-core setting to filter low-frequency nodes. To align items to knowledge graph entities, we adopt the same procedure as ~\cite{kgcn,wang2019kgat}.

\begin{table}
\caption{Statistics of HGB datasets.}\label{tab:statistics}
\small
\setlength{\tabcolsep}{0.9mm}
\begin{tabular}{crcrccc}
\toprule
 \makecell[c]{\emph{Node}\\\emph{Classification}} & \#Nodes & \makecell[c]{\#Node\\Types} & \#Edges & \makecell[c]{\#Edge\\Types}& Target& \#Classes\\ 
 \midrule
DBLP          & 26,128               & 4                         & 239,566              & 6                         & author                                                                             & 4                            \\
IMDB          & 21,420               & 4                         & 86,642               & 6                         & movie                                                                              & 5                            \\
ACM           & 10,942               & 4                         & 547,872              & 8                         & paper                                                                              & 3                            \\
Freebase      & 180,098              & 8                         & 1,057,688            & 36                        & book                                                                               & 7                            \\
\midrule
\emph{Link Prediction}& & & & & \multicolumn{2}{c}{Target}\\
\midrule
Amazon        & 10,099               & 1                         & 148,659              & 2                         & \multicolumn{2}{c}{product-product}                            \\
LastFM        & 20,612               & 3                         & 141,521              & 3                         & \multicolumn{2}{c}{user-artist}                           \\
PubMed        & 63,109               & 4                         & 244,986              & 10                        & \multicolumn{2}{c}{disease-disease} \\
\midrule
\midrule
\end{tabular}
 \begin{tabular}{crrrrl}
            \emph{Recommendation}    & \multicolumn{1}{c}{Amazon-book} & \multicolumn{1}{c}{LastFM} & \multicolumn{1}{c}{Movielens} & \multicolumn{1}{c}{Yelp-2018} &  \\ \midrule
\#Users      & 70,679                          & 23,566                     & 37,385                        & 45,919                        &  \\
\#Items        & 24,915                          & 48,123                     & 6,182                         & 45,538                        &  \\
\#Interactions & 846,434                         & 3,034,763                  & 539,300                       & 1,183,610                     &  \\
\#Entities     & 113,487                         & 106,389                    & 24,536                        & 136,499                       &  \\
\#Relations    & 39                              & 9                          & 20                            & 42                            &  \\
\#Triplets     & 2,557,746                       & 464,567                    & 237,155                       & 1,853,704                     &  \\ \bottomrule
\end{tabular}

\end{table}

\subsection{Feature Preprocessing}\label{sec:input}

As pointed out in \S~\ref{sec:motivation}, the preprocessing for input features has a great impact on the performance. Our preprocessing methods are as follows.

\vpara{Linear Transformation.} As the input feature of different types of nodes may vary in dimension, we use a linear layer with bias for each node type to map all node features to a shared feature space. The parameters in these linear layers will be optimized along with the following HGNN. 

\vpara{Useful Types Selection.} In many datasets, only features of a part of types are useful to the task. We can select a subset of node types to keep their features, and replace the features of nodes of other types as one-hot vectors. Combined with linear transformation, the replacement is equivalent to learn an individual embedding for each node of the unselected types. Ideally, we should enumerate all subsets of types and report the best one based on the performance on the validation set, but due to the high consumption to train the model $2^{|T_v|}$ times, we decide to only enumerate three choices, i.e. using all given node features, using only features of target node type, or replacing all node features as one-hot vectors.

\subsection{Downstream Decoders and Loss function}


\subsubsection{Node Classification}
After setting the final dimension of HGNNs the same as the number of classes, we then adopt the most usual loss functions.
For single-label classification, we use softmax and cross-entropy loss. For multi-label datasets, i.e. IMDB in HGB, we use a sigmoid activation and binary cross-entropy loss.

\subsubsection{Link Prediction}

As RGCN~\cite{battaglia2018relational} suggests, DistMult~\cite{yang2014embedding} performs better than direct dot product, due to multiple types of edges, i.e. for node pair $u,v$ and a target edge type $r$,
\begin{equation}
    \text{Prob}_r(u,v \text{ is positive}) = \text{sigmoid}\left(\text{HGNN}(u)^T R_r \ \text{HGNN}(v)\right),
\end{equation}
where $R_r$ is a learnable square matrix (sometimes regularized with diagonal matrix) for type $r\in T_e$. We find that DistMult outperforms dot product sometimes even when there is only single type of edge to predict.
We try both dot product and DistMult decoders, and report the best results. The loss function is binary cross-entropy.

\subsubsection{Knowledge-aware Recommendation}
Recommendation is similar to  link prediction, but differs in data distribution and focuses more on ranking. We define the similarity function $f(u,v)$ between nodes $u,v$ based on dot product. As mentioned in \S~\ref{sec:kgat}, pretrained BPR-MF embeddings are of vital importance. We incorporate the BPR-MF embeddings $e_u, e_v$ via a bias term in $f(u,v)$ to avoid modification on the input or architectures of other models, i.e.,
\begin{equation}
    f(u,v) = \text{HGNN}(u)^T\text{HGNN}(v) + e_u^T\ e_v.
\end{equation}
Following KGAT~\cite{wang2019kgat}, we opt for BPR~\cite{rendle2009bpr} loss for training.
\begin{equation}
    \text{Loss}(u, v^+, v^-)= -\log \text{sigmoid}\left(f(u,v^+) - f(u, v^-)\right),
\end{equation}
where $(u, v^+)$ is a positive pair, and $(u, v^-)$ is a negative pair sampled at random.





\begin{figure*}
    \centering
    \includegraphics[width=\textwidth]{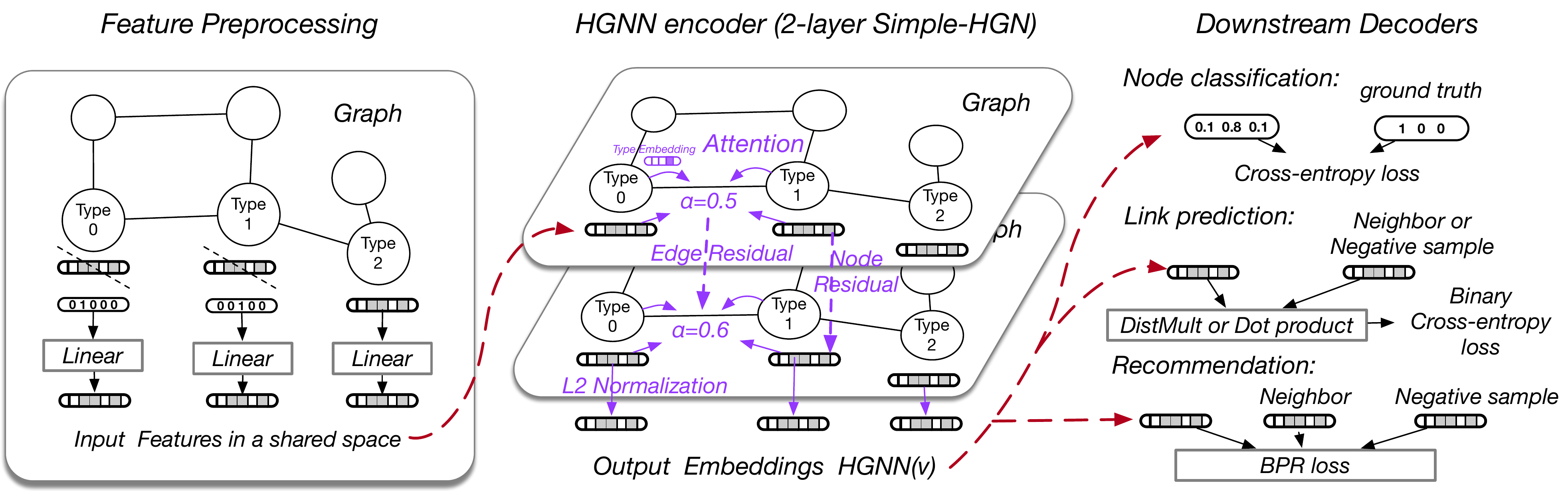}
    \caption{HGB pipeline and \model. { In this illustration, we assume only the features of Type 2 nodes are kept in the \emph{Feature Preprocessing} period. The purple parts are the improvements over GAT in \model.}}
    \label{fig:main}
\end{figure*}

\subsection{Evaluation Settings}

We evaluate all methods for all datasets by running 5 times with different random seeds, and reporting the average score and standard deviation.

\subsubsection{Node Classification} We evaluate node classification with Macro-F1 and Micro-F1 metrics for both multi-class (DBLP, ACM, Freebase) and multi-label (IMDB) datasets. The implementation is based on sklearn\footnote{https://scikit-learn.org/stable/modules/generated/sklearn.metrics.f1\_score.html}. 

\subsubsection{Link Prediction} We evaluate link prediction with ROC-AUC (area under the ROC curve) and MRR (mean reciprocal rank) metrics. Since we usually need to determine a threshold when to classify a pair as positive for the score given by decoder, ROC-AUC can evaluate the performance under difference threshold at a whole scope. MRR can evaluate the ranking performance for different methods. Following~\cite{yang2020heterogeneous}, we calculate MRR scores clustered by the head of pairs in test set, and return the average of them as MRR performance.

\subsubsection{Knowledge-aware Recommendation}
Recommendation task focuses more on ranking instead of classification. Therefore, we adopt recall@20 and ndcg@20 as our evaluation metrics. 
The average metrics for all users in test set are reported as benchmark performance.

\section{A Simple Heterogeneous GNN}\label{sec:model}


Inspired by the advantage of the simple GAT over advanced and dedicated heterogeneous GNNs, 
we present \model, a simple and effective method for modeling heterogeneous graph.  
{\smodel adopts GAT as backbone} 
with enhancements from the redesign of three well-known techniques: 
Learnable edge-type embedding, residual connections, and $L_2$ normalization on the output embeddings.
Figure~\ref{fig:main} illustrates the full pipeline with \smodel.




\subsection{Learnable Edge-type Embedding} Though GAT has powerful capacity in modeling homogeneous graphs, it may be not optimal for heterogeneous graphs due to the neglect of node or edge types.  To tackle this problem, we extend the original graph attention mechanism by including edge type information into attention calculation.  Specifically, at each layer, we allocate a $d_l-$dimensional embedding $\*r^{(l)}_{\psi{(e)}}$ for each edge type $\psi(e) \in T_e$, and use both edge type embeddings and node embeddings to calculate the attention score as follows\footnote{We omit the superscript ${(l)}$ in this equation for the sake of brevity.}:
\begin{equation}
\label{eq:ratt}
\hat{\alpha}_{ij} = \frac{\exp\left(\text{LeakyReLU}\left(\*a^T[\*W\*h_i\|\*W\*h_j
\textcolor{purple2}{\|\*W_r\*r_{\psi{(\langle i,j\rangle )}}}
]\right)\right)}{\sum_{k\in \&N_i} \exp\left(\text{LeakyReLU}\left(\*a^T[\*W\*h_i\|\*W\*h_k\textcolor{purple2}{\|\*W_r\*r_{\psi{(\langle i,k\rangle )}}}
]\right)\right)},
\end{equation}
where $\psi{(\langle i,j\rangle )}$ represents the type of edge between node $i$ and node $j$, and $\*W_r^{(l)}$ is a learnable matrix to transform type embeddings.

\begin{table*}[ht]
\caption{Node classification benchmark. Vacant positions (``-'') mean that the models run out of memory on large graphs.}\label{tab:NC}
\small
\begin{tabular}{ccccccccc}
\toprule
                    & \multicolumn{2}{c}{DBLP}                      & \multicolumn{2}{c}{IMDB}                      & \multicolumn{2}{c}{ACM}                       & \multicolumn{2}{c}{Freebase}                   \\ \midrule
                    & Macro-F1              & Micro-F1              & Macro-F1              & Micro-F1              & Macro-F1              & Micro-F1              & Macro-F1              & Micro-F1               \\ \midrule
RGCN	&	91.52$\pm$0.50	&	92.07$\pm$0.50	&	58.85$\pm$0.26	&	62.05$\pm$0.15	&	91.55$\pm$0.74	&	91.41$\pm$0.75	&	46.78$\pm$0.77	&	58.33$\pm$1.57\\
HAN	&	91.67$\pm$0.49	&	92.05$\pm$0.62	&	57.74$\pm$0.96	&	64.63$\pm$0.58	&	90.89$\pm$0.43	&	90.79$\pm$0.43	&	21.31$\pm$1.68	&	54.77$\pm$1.40\\
GTN	&	93.52$\pm$0.55	&	93.97$\pm$0.54	&	60.47$\pm$0.98	&	65.14$\pm$0.45	&	91.31$\pm$0.70	&	91.20$\pm$0.71	&	-	&	-\\
RSHN	&	93.34$\pm$0.58	&	93.81$\pm$0.55	&	59.85$\pm$3.21	&	64.22$\pm$1.03	&	90.50$\pm$1.51	&	90.32$\pm$1.54	&	-	&	-\\
HetGNN	&	91.76$\pm$0.43	&	92.33$\pm$0.41	&	48.25$\pm$0.67	&	51.16$\pm$0.65	&	85.91$\pm$0.25	&	86.05$\pm$0.25	&	-	&	-\\
MAGNN	&	93.28$\pm$0.51	&	93.76$\pm$0.45	&	56.49$\pm$3.20	&	64.67$\pm$1.67	&	90.88$\pm$0.64	&	90.77$\pm$0.65	&	-	&	-\\
HetSANN	&	78.55$\pm$2.42	&	80.56$\pm$1.50	&	49.47$\pm$1.21	&	57.68$\pm$0.44	&	90.02$\pm$0.35	&	89.91$\pm$0.37	&	-	&	-\\
HGT	&	93.01$\pm$0.23	&	93.49$\pm$0.25	&	63.00$\pm$1.19	&	67.20$\pm$0.57	&	91.12$\pm$0.76	&	91.00$\pm$0.76	&	29.28$\pm$2.52	&	60.51$\pm$1.16\\ \midrule
GCN	&	90.84$\pm$0.32	&	91.47$\pm$0.34	&	57.88$\pm$1.18	&	64.82$\pm$0.64	&	92.17$\pm$0.24	&	92.12$\pm$0.23	&	27.84$\pm$3.13	&	60.23$\pm$0.92\\
GAT	&	93.83$\pm$0.27	&	93.39$\pm$0.30	&	58.94$\pm$1.35	&	64.86$\pm$0.43	&	92.26$\pm$0.94	&	92.19$\pm$0.93	&	40.74$\pm$2.58	&	65.26$\pm$0.80\\ \midrule
Simple-HGN & \textbf{94.01$\pm$0.24} & \textbf{94.46$\pm$0.22} & \textbf{63.53$\pm$1.36} & \textbf{67.36$\pm$0.57} & \textbf{93.42$\pm$0.44} & \textbf{93.35$\pm$0.45} & \textbf{47.72$\pm$1.48} & \textbf{66.29$\pm$0.45}  \\
\bottomrule
\end{tabular}
\end{table*}

\begin{table*}[ht]
\caption{Link prediction benchmark. Vacant positions (``-'') are due to lack of meta-paths on those datasets.}\label{tab:LP}
\small
\begin{tabular}{ccccccc}
\toprule
           & \multicolumn{2}{c}{Amazon}                   & \multicolumn{2}{c}{LastFM}                   & \multicolumn{2}{c}{PubMed}                     \\ \midrule
           & ROC-AUC               & MRR                  & ROC-AUC               & MRR                   & ROC-AUC               & MRR                    \\ \midrule
RGCN	&	86.34$\pm$0.28	&	93.92$\pm$0.16	&	57.21$\pm$0.09	&	77.68$\pm$0.17	&	78.29$\pm$0.18	&	90.26$\pm$0.24\\
GATNE	&	77.39$\pm$0.50	&	92.04$\pm$0.36	&	66.87$\pm$0.16	&	85.93$\pm$0.63	&	63.39$\pm$0.65	&	80.05$\pm$0.22\\
HetGNN	&	77.74$\pm$0.24	&	91.79$\pm$0.03	&	62.09$\pm$0.01	&	83.56$\pm$0.14	&	73.63$\pm$0.01	&	84.00$\pm$0.04\\
MAGNN	&	-	&	-	&	56.81$\pm$0.05	&	72.93$\pm$0.59	&	-	&	-\\
HGT	&	88.26$\pm$2.06	&	93.87$\pm$0.65	&	54.99$\pm$0.28	&	74.96$\pm$1.46	&	80.12$\pm$0.93	&	90.85$\pm$0.33\\ \midrule
GCN	&	92.84$\pm$0.34	&	\textbf{97.05$\pm$0.12}	&	59.17$\pm$0.31	&	79.38$\pm$0.65	&	80.48$\pm$0.81	&	90.99$\pm$0.56\\
GAT	&	91.65$\pm$0.80	&	96.58$\pm$0.26	&	58.56$\pm$0.66	&	77.04$\pm$2.11	&	78.05$\pm$1.77	&	90.02$\pm$0.53\\ \midrule
Simple-HGN & \textbf{93.40$\pm$0.62} & 96.94$\pm$0.29         & \textbf{67.59$\pm$0.23} & \textbf{90.81$\pm$0.32} & \textbf{83.39$\pm$0.39} & \textbf{92.07$\pm$0.26} \\
\bottomrule
\end{tabular}
\end{table*}

\begin{table*}[ht]
\caption{Knowledge-aware recommendation benchmark. KGAT- refers to KGAT without redundant designs~(See \S~\ref{sec:kgat}). GCN and GAT are not included, because they are already very similar to KGCN and KGAT-. The KGCN and KGAT works focus more on incorporating knowledge into user-item graphs than new architectures.}
\label{tab:recom_benchmark}
\centering
\footnotesize
\begin{tabular}{ccccccccc}
\toprule
        & \multicolumn{2}{c}{Amazon-Book}   & \multicolumn{2}{c}{LastFM} & \multicolumn{2}{c}{Yelp-2018}     & \multicolumn{2}{c}{MovieLens}  \\ \midrule
        & recall@20       & ndcg@20         & recall@20       & ndcg@20 & recall@20       & ndcg@20         & recall@20       & ndcg@20\\ \midrule
KGCN	&	0.1464$\pm$0.0002	&	0.0769$\pm$0.0002	&	0.0819$\pm$0.0002	&	0.0705$\pm$0.0002	&	0.0683$\pm$0.0003	&	0.0431$\pm$0.0003	&	0.4237$\pm$0.0008	&	0.2753$\pm$0.0005\\
KGNN-LS	&	0.1448$\pm$0.0003	&	0.0759$\pm$0.0001	&	0.0806$\pm$0.0003	&	0.0695$\pm$0.0002	&	0.0671$\pm$0.0003	&	0.0422$\pm$0.0002	&	0.4218$\pm$0.0008	&	0.2741$\pm$0.0005\\
KGAT	&	0.1507$\pm$0.0003	&	0.0802$\pm$0.0004	&	0.0877$\pm$0.0003	&	0.0749$\pm$0.0003	&	0.0697$\pm$0.0002	&	0.0450$\pm$0.0001	&	0.4532$\pm$0.0004	&	0.3007$\pm$0.0008\\
KGAT$-$	&	0.1486$\pm$0.0003	&	0.0790$\pm$0.0002	&	0.0890$\pm$0.0002	&	0.0762$\pm$0.0002	&	0.0715$\pm$0.0001	&	0.0460$\pm$0.0001	&	0.4553$\pm$0.0003	&	0.3031$\pm$0.0006\\ \midrule
    Simple-HGN	&	\textbf{0.1587$\pm$0.0011}	&	\textbf{0.0854$\pm$0.0005}	&	\textbf{0.0917$\pm$0.0006}	&	\textbf{0.0797$\pm$0.0003}	&	\textbf{0.0732$\pm$0.0003}	&	\textbf{0.0466$\pm$0.0003}	&	\textbf{0.4618$\pm$0.0007}	&	\textbf{0.3090$\pm$0.0007}\\

\bottomrule
\end{tabular}
\end{table*}

\subsection{Residual Connection} GNNs are hard to be deep due to the over-smoothing and gradient vanishing problems~\cite{li2018deeper,xu2018representation}. A famous solution to mitigate this problem in computer vision is residual connection~\cite{resnet}. However, the original GCN paper~\cite{gcn} showed a negative result for residual connection on graph convolution. Recent study~\cite{li2020deepergcn} finds that well-designed pre-activation implementation could make residual connection great again in GNNs.

\vpara{Node Residual.} We add pre-activation residual connection for node representation across layers. The aggregation at the $l^{th}$ layer can be expressed as 
\begin{equation}
\label{eq:singleatt}
     \*h^{(l)}_i =  \sigma\left(\sum_{j\in\mathcal{N}_i}{ \alpha^{(l)}_{ij}\*W^{(l)}\*h^{(l-1)}_j} +  \*h^{(l-1)}_i\right),
\end{equation}
where $\alpha^{(l)}_{ij}$ is the attention weight about edge $\langle i,j\rangle $ and $\sigma$ is an activation function (ELU~\cite{clevert2015fast} by default). When the dimension changes in the $l-$th layer, an additional learnable linear transformation $\*W^{(l)}_{res}\in \mathbb{R}^{d_{l+1} \times d_l}$ is needed, i.e.,
\begin{equation}
\label{eq:singleatt2}
     \*h^{(l)}_i =  \sigma\left(\sum_{j\in\mathcal{N}_i}{ \alpha^{(l)}_{ij}\*W^{(l)}\*h^{(l-1)}_j} +  \*W^{(l)}_{\text{res}}\*h^{(l-1)}_i\right). 
\end{equation}
\vpara{Edge Residual.} Recently, Realformer~\cite{realformer} reveals that residual connection on attention scores is also helpful. After getting the raw attention scores $\hat{\alpha}$ via Eq.~\eqref{eq:ratt}, we add residual connections to them,
\begin{equation}
\label{eq:attres}
    \alpha^{(l)}_{ij} = (1-\beta) \hat{\alpha}^{(l)}_{ij} + \beta \alpha^{(l-1)}_{ij},
\end{equation}
where hyperparameter $\beta \in [0,1]$ is a scaling factor.



\vpara{Multi-head Attention.} Similar to GAT, we adopt multi-head attention to enhance model's expressive capacity. Specifically, we perform $K$ independent attention mechanisms according to Equation~\eqref{eq:singleatt}, and concatenate their results as the final representation. The corresponding updating rule is:
\begin{equation}
\label{eq:multiattres}
    \alpha^{(l)}_{ijk} = (1-\beta) \hat{\alpha}^{(l)}_{ijk} + \beta \alpha^{(l-1)}_{ijk},
\end{equation}
\begin{equation}
\label{eq:multihead}
     \hat{\*h}^{(l)}_{ik} =  \sum_{j\in\mathcal{N}_i}{ \alpha^{(l)}_{ijk}\*W_k^{(l)}\*h^{(l-1)}_j},
\end{equation}
\begin{equation}
\label{eq:multihead}
     \*h^{(l)}_i =  \sigma\left(\concat_{k=1}^{K}{\hat{\*h}_{ik}^{(l)}} + \*W^{(l)}_{\text{res}(k)} \*h^{(l-1)}_i\right),
\end{equation}
where $\concat$ denotes concatenation operation, and $\hat{\alpha}^{(l)}_{ijk}$ is attention score computed by the $k^{th}$ linear transformation $\*W^{l}_{k}$ according to Equation~\eqref{eq:singleatt2}.

Usually the output dimension cannot be divided exactly by the number of heads. Following GAT, we no longer use concatenation but adopt averaging for the representation in the final ($L^{th}$) layer, i.e.,
\begin{equation}
\label{eq:l2}
    \*h_i^{(L)} = \frac{1}{K}
    \sum_{k=1}^{K} \hat{\*h}^{(L)}_{ik}.
\end{equation}

\vpara{Adaptation for Link Prediction.} We slightly modify the model architecture for better performance on link prediction. Edge residual is removed and the final embedding is the concatenation of embeddings from all the layers. This adapted version is similar to JKNet~\cite{xu2018representation}.

\begin{table*}[t]
\caption{Ablation studies for \model.}
\footnotesize
\label{tab:ablation}
\begin{tabular}{ccccccc}
\toprule
Task                      & \multicolumn{2}{c}{Node Classification} & \multicolumn{2}{c}{Link Prediction} & \multicolumn{2}{c}{Recommendation}   \\ \midrule
Dataset                   & \multicolumn{2}{c}{IMDB}                & \multicolumn{2}{c}{Last.fm}         & \multicolumn{2}{c}{Movielens-20M}    \\ \midrule
  Metric                        & Macro-F1        & Micro-F1              & ROC-AUC         & MRR               & recall@20        & ndcg@20           \\ \midrule
Simple-HGN                & \textbf{63.53$\pm$1.36} & \textbf{67.36$\pm$0.57}       & 67.59$\pm$0.23           & \textbf{90.81$\pm$0.32}   & 0.4626$\pm$0.0006           & 0.3532$\pm$0.0005            \\
w.o. type embedding   & 63.04$\pm$1.00           & 67.06$\pm$0.40                 & \textbf{67.61$\pm$0.13} & 90.52$\pm$0.13             & \textbf{0.4632$\pm$0.0005} & \textbf{0.3537$\pm$0.0007}  \\
w.o. L2 normalization     & 58.06$\pm$1.62           & 65.33$\pm$0.69                 & 61.07$\pm$0.96           & 82.51$\pm$1.56             & 0.3837$\pm$0.0187           & 0.2816$\pm$0.0173            \\
w.o. residual connections & 61.61$\pm$2.34           & 66.28$\pm$1.11                 & 63.33$\pm$0.78           & 84.13$\pm$0.62             & 0.4261$\pm$0.0004           & 0.3192$\pm$0.0006           \\ \bottomrule
\end{tabular}
\end{table*}

\subsection{$L_2$ Normalization} We find that an $L_2$ normalization on the output embedding is extremely useful, i.e., 
\begin{equation}
\label{eq:l2_2}
    \*o_i = \frac{\*h_i^{(L)}}{\lVert \*h_i^{(L)} \rVert},
\end{equation}
where $\*o_i$ is the output embedding of node $i$ and $\*h_i^{(L)}$ is the final representation from Equation~\eqref{eq:l2}.

The normalization on the output embedding is very common for retrieval-based tasks, because the dot product will be equivalent to the cosine similarity after normalization. But we also find its improvements for classification tasks, which was also observed in computer vision~\cite{l2norm}. 
Additionally, it suggests to multiply a scaling parameter $\alpha$ to the output embedding~\cite{l2norm}. 
We find that tuning an appropriate $\alpha$ indeed improves the performance, but it varies a lot in different datasets. 
We thus keep the form of Eq.~\eqref{eq:l2_2} for simplicity.

\hide{

\section{Simple Heterogeneous Graph Neural Network}\label{sec:model}
\begin{figure*}
    \centering
    \includegraphics[width=\textwidth]{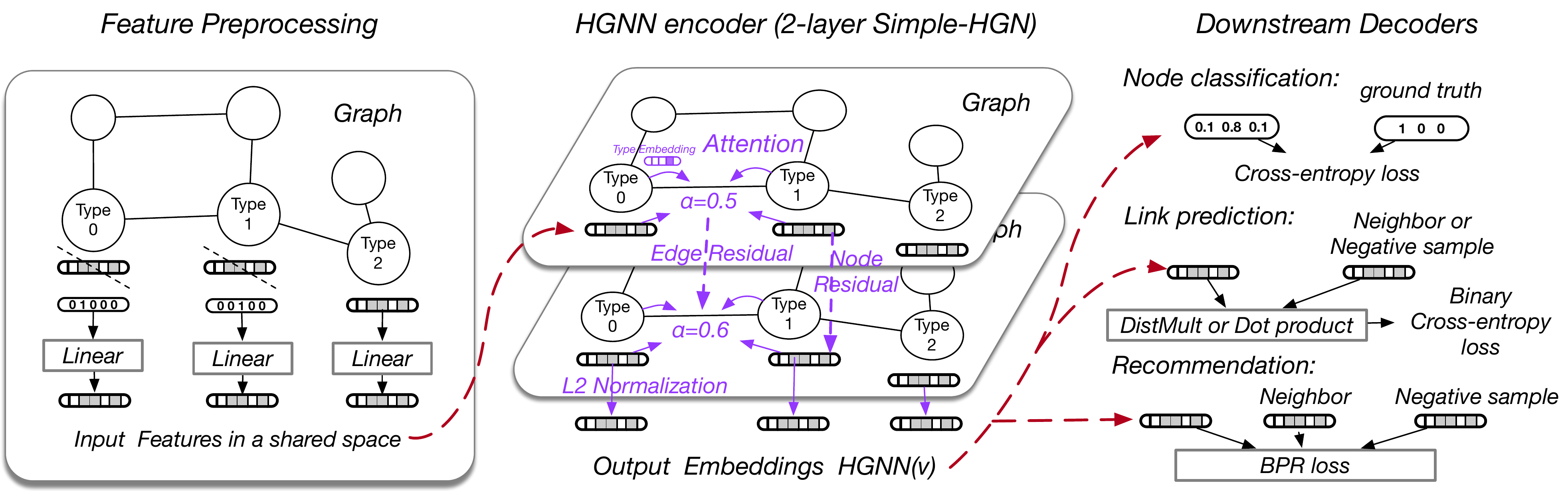}
    \caption{HGB pipeline and \model. { In this illustration, we assume only the features of Type 2 nodes are kept in the \emph{Feature Preprocessing} period. The purple parts are the improvements over GAT in \model.}}
    \label{fig:main}
\end{figure*}
To investigate the room for improvement of HGB under the current GNN techniques, and setup a strong baseline for future works, 
we propose \model , a simple and effective method for modeling heterogeneous graph.  \textbf{\smodel adopts GAT as backbone} and enhances it with three modifications: Learnable edge-type embedding, residual connections and $L_2$ normalization on the output embeddings.

The three techniques, especially the latter two, are found from a great quantity of ablation studies.  Many techniques in GNNs and related domains are tested on HGB, and the residual connections and $L_2$ normalization finally emerged from them with steady improvement and simplicity. We detail them in this section and illustrate the full pipeline with \smodel in Figure~\ref{fig:main}.


\subsection{Learnable Edge-type Embedding} Though GAT has powerful capacity in modeling homogeneous graph, it may be not optimal for heterogeneous graph due to the neglect of node or edge types.  To tackle this problem, we extend the original graph attention mechanism by including edge type information into attention calculation.  Specifically, at each layer, we allocate a $d_l-$dimensional embedding $\*r^{(l)}_{\psi{(e)}}$ for each edge type $\psi(e) \in T_e$, and use both both edge type embeddings and node embeddings to calculate the attention score as follows\footnote{We omit the superscript ${(l)}$ in this equation for the sake of brevity.}:
\begin{equation}
\label{eq:ratt}
\hat{\alpha}_{ij} = \frac{\exp\left(\text{LeakyReLU}\left(\*a^T[\*W\*h_i\|\*W\*h_j
\textcolor{blue}{\|\*W_r\*r_{\psi{(\langle i,j\rangle )}}}
]\right)\right)}{\sum_{k\in \&N_i} \exp\left(\text{LeakyReLU}\left(\*a^T[\*W\*h_i\|\*W\*h_k\textcolor{blue}{\|\*W_r\*r_{\psi{(\langle i,k\rangle )}}}
]\right)\right)},
\end{equation}
where $\psi{(\langle i,j\rangle )}$ represents the type of edge between node $i$ and node $j$.

\subsection{Residual Connection} GNNs are hard to be deep due to the over-smoothing and gradient vanishing problems~\cite{li2018deeper,xu2018representation}. A famous solution to mitigate this problem in computer vision is residual connection~\cite{resnet}. However, the original GCN paper~\cite{gcn} showed a negative result for residual connection on graph convolution. Recent study~\cite{li2020deepergcn} finds that well-designed pre-activation implementation could make residual connection great again in GNNs.

\vpara{Node Residual.} We add pre-activation residual connection for node representation across layers. The aggregation at the $l^{th}$ layer can be expressed as 
\begin{equation}
\label{eq:singleatt}
     \*h^{(l)}_i =  \sigma\left(\sum_{j\in\mathcal{N}_i}{ \alpha^{(l)}_{ij}\*W^{(l)}\*h^{(l-1)}_j} +  \*h^{(l-1)}_i\right),
\end{equation}
where $\alpha^{(l)}_{ij}$ is the attention weight about edge $\langle i,j\rangle $ and $\sigma$ is an activation function (ELU~\cite{clevert2015fast} by default). When the dimension changes in the $l-$th layer, an additional learnable linear transformation $\*W^{(l)}_{res}\in \mathbb{R}^{d_{l+1} \times d_l}$ is needed, i.e.,
\begin{equation}
\label{eq:singleatt2}
     \*h^{(l)}_i =  \sigma\left(\sum_{j\in\mathcal{N}_i}{ \alpha^{(l)}_{ij}\*W^{(l)}\*h^{(l-1)}_j} +  \*W^{(l)}_{\text{res}}\*h^{(l-1)}_i\right). 
\end{equation}
\vpara{Edge Residual.} Recently, Realformer~\cite{realformer} reveals that residual connection on attention scores is also helpful. After getting the raw attention scores $\hat{\alpha}$ via Equation~\eqref{eq:ratt}, we add residual connections to them,
\begin{equation}
\label{eq:attres}
    \alpha^{(l)}_{ij} = (1-\beta) \hat{\alpha}^{(l)}_{ij} + \beta \alpha^{(l-1)}_{ij},
\end{equation}
where hyperparameter $\beta \in [0,1]$ is a scaling factor.



\vpara{Multi-head Attention.} Similar to GAT, we adopt multi-head attention to enhance model's expressive capacity. Specifically, we perform $K$ independent attention mechanisms according to Equation~\eqref{eq:singleatt}, and concatenate their results as the final representation. The corresponding updating rule is:
\begin{equation}
\label{eq:multiattres}
    \alpha^{(l)}_{ijk} = (1-\beta) \hat{\alpha}^{(l)}_{ijk} + \beta \alpha^{(l-1)}_{ijk},
\end{equation}
\begin{equation}
\label{eq:multihead}
     \hat{\*h}^{(l)}_{ik} =  \sum_{j\in\mathcal{N}_i}{ \alpha^{(l)}_{ijk}\*W_k^{(l)}\*h^{(l-1)}_j},
\end{equation}
\begin{equation}
\label{eq:multihead}
     \*h^{(l)}_i =  \sigma\left(\concat_{k=1}^{K}{\hat{\*h}_{ik}^{(l)}} + \*W^{(l)}_{\text{res}(k)} \*h^{(l-1)}_i\right),
\end{equation}
where $\concat$ denotes concatenation operation, and $\hat{\alpha}^{(l)}_{ijk}$ is attention score computed by the $k^{th}$ linear transformation $\*W^{l}_{k}$ according to Equation~\eqref{eq:singleatt2}.

Usually the output dimension cannot be divided exactly by the number of heads. Following GAT, we no longer use concatenation but adopt averaging for the representation in the final ($L^{th}$) layer, i.e.,
\begin{equation}
\label{eq:l2}
    \*h_i^{(L)} = \frac{1}{K}
    \sum_{k=1}^{K} \hat{\*h}^{(L)}_{ik}.
\end{equation}

\vpara{Adaptation for Link Prediction.} We slightly modify the model architecture for better performance on link prediction. Edge residual is removed and the final embedding is the concatenation of embeddings from all the layers. This adapted version is similar to JKNet~\cite{xu2018representation}.

\begin{table*}[t]
\caption{Ablation studies for \model.}
\label{tab:ablation}
\begin{tabular}{ccccccc}
\toprule
Task                      & \multicolumn{2}{c}{Node Classification} & \multicolumn{2}{c}{Link Prediction} & \multicolumn{2}{c}{Recommendation}   \\ \midrule
Dataset                   & \multicolumn{2}{c}{IMDB}                & \multicolumn{2}{c}{Last.fm}         & \multicolumn{2}{c}{Movielens-20M}    \\ \midrule
  Metric                        & Macro-F1        & Micro-F1              & ROC-AUC         & MRR               & recall@20        & ndcg@20           \\ \midrule
Simple-HGN                & \textbf{63.53$\pm$1.36} & \textbf{67.36$\pm$0.57}       & 67.59$\pm$0.23           & \textbf{90.81$\pm$0.32}   & 0.4626$\pm$0.0006           & 0.3532$\pm$0.0005            \\
w.o. type embedding   & 63.04$\pm$1.00           & 67.06$\pm$0.40                 & \textbf{67.61$\pm$0.13} & 90.52$\pm$0.13             & \textbf{0.4632$\pm$0.0005} & \textbf{0.3537$\pm$0.0007}  \\
w.o. L2 normalization     & 58.06$\pm$1.62           & 65.33$\pm$0.69                 & 61.07$\pm$0.96           & 82.51$\pm$1.56             & 0.3837$\pm$0.0187           & 0.2816$\pm$0.0173            \\
w.o. residual connections & 61.61$\pm$2.34           & 66.28$\pm$1.11                 & 63.33$\pm$0.78           & 84.13$\pm$0.62             & 0.4261$\pm$0.0004           & 0.3192$\pm$0.0006           \\ \bottomrule
\end{tabular}
\end{table*}

\subsection{$L_2$ Normalization} We find that an $L_2$ normalization on the output embedding is extremely useful, i.e., 
\begin{equation}
\label{eq:l2_2}
    \*o_i = \frac{\*h_i^{(L)}}{\lVert \*h_i^{(L)} \rVert},
\end{equation}
where $\*o_i$ is the output embedding of node $i$ and $\*h_i^{(L)}$ is the final representation from Equation~\eqref{eq:l2}.

The normalization on the output embedding is very common for retrieval-based tasks, because the dot product will be equivalent to the cosine similarity after normalization. But we also find an improvement in classification tasks, which has been observed in computer vision domain~\cite{l2norm}. That paper~\cite{l2norm} also suggests to multiply a scaling parameter $\alpha$ to the output embedding. We find that an appropriate $\alpha$ indeed improves the performance, but it varies a lot in different datasets. We decide to keep the form of Equation~\eqref{eq:l2_2} for simplicity.

}



\hide{
\begin{equation}
\label{eq:concat}
    \mathbf{h}^{(l)}_i = \concat_{k=1}^{T} \mathbf{o}^{k(l)}_i
\end{equation}

\begin{equation}
\label{eq:l2}
    \mathbf{h}'^{(l)}_i  = \frac{1}{T}
    \sum_{k=1}^{T} \mathbf{o}^{k(l)}_i
\end{equation}

\begin{equation}
\label{eq:l2_2}
    \mathbf{h}^{(l)}_i = \frac{\mathbf{h}'^{(l)}_i}{\lVert \mathbf{h}'^{(l)}_i \rVert}
\end{equation}
}
\hide{
In order to encode heterogeneous graph information, we propose a rather simple baseline method, but performs astonishingly well. Concretely, we first use a node-type specific linear transformation to map heterogeneous node features into a unified feature space. Then, we stack multiple relation-aware graph attention layers to aggregate information from heterogeneous neighbors. After performing residual connection and row-wise L2 normalization on the graph layers, the final node embeddings are obtained.

Here is a formalized description of our method. Suppose the input features of nodes are $\mathbf{H}^{(0)}=\{\mathbf{H}^{(0)}_1, \mathbf{H}^{(0)}_2, ..., \mathbf{H}^{(0)}_K\}$, where $\mathbf{H}^{(0)}_i \in \mathbb{R}^{N_i\times F_i}$, $K$ is the number of node types, $N_i$ is the number of nodes for node type $i$, and $F_i$ is the dimension of node features for node type $i$.

After taking linear transformations, the node features become $\mathbf{H}^{(1)}=\{\mathbf{H}^{(1)}_1, \mathbf{H}^{(1)}_2, ..., \mathbf{H}^{(1)}_K\}$, in which $\mathbf{H}^{(1)}_i = \mathbf{H}^{(0)}_i \mathbf{W}^{(0)}_i+\mathbf{b}^{(0)}_i$, where $\mathbf{W}^{(0)}_i \in \mathbb{R}^{F_i \times F}$ and $\mathbf{b}^{(0)}_i \in \mathbb{R}^{1 \times F}$ are trainable parameters. Now we can see $\mathbf{H}^{(1)}$ as a $\mathbb{R}^{N\times F}$ matrix and feed it to GNN layers, where $N=\sum_{i=1}^K N_i$ and $F$ is the dimension of unified feature space.

To tackle with heterogeneous graph information, we propose relation-aware graph attention layer, extending vanilla graph attention layer to heterogeneous domain. Specifically, we allocate a $d$ dimensional embedding for each relation type. When calculating attention value for a given pair of nodes, we not only concatenate the embedding of the pair of nodes, but also the corresponding relation type embedding. Moreover, motivated by ResNet~\cite{resnet} and RealFormer~\cite{realformer}, we also add optional residual connection for node embeddings and attention scores for these graph layers. Formally,

\begin{equation}
    \alpha^{(l)}_{ij} = (1-\beta) \hat{\alpha}^{(l)}_{ij} + \beta \alpha^{(l-1)}_{ij}
\end{equation}

\begin{equation}
    \mathbf{o}^{(l+1)}_i =  \sum_{j\in\mathcal{N}_i}{ \alpha^{(l)}_{i,j}\mathbf{h}^{(l)}_j \mathbf{W}}
\end{equation}

To adapt multi-head attention, we repeat the above attention mechanism $T$ times, and aggregate these embeddings to get the final representation.

\begin{equation}
    \mathbf{h}'^{(l+1)}_i = \text{AGGREGATE}(\{\mathbf{o}^{k(l+1)}_i, \forall k\in [1..T]\})
\end{equation}

where $\text{AGGREGATE}$ is one of some pre-defined aggregation operations.

\begin{equation}
    \mathbf{h}^{(l+1)}_i = \sigma \left(\mathbf{h}'^{(l+1)}_i + \mathbf{h}^{(l)}_i \mathbf{W}_{\text{res}}\right)
\end{equation}

where $\mathbf{h}_i^{(l)}$ is the $i$-th row of $\mathbf{H}^{(l)}$, $\mathbf{o}_i^{(l)}$ is the $i$-th row of a single head output matrix $\mathbf{O}^{(l)}$, and $\mathcal{N}_i$ is the set of node $i$'s neighbors. $\mathbf{W}$, $\mathbf{W}_r$ and $\mathbf{a}^T$ are trainable parameters. $\sigma$ is activation function, where we use exponential linear unit (ELU) by default. For residual connection, $\mathbf{W}_{\text{res}}$ is identity matrix if it connects two vector with same dimension, otherwise, it is a trainable linear transformation. If residual connection is not used, $\mathbf{W}_{\text{res}}$ is a zero matrix. For residual attention, $\beta \in [0,1]$ is a hyper-parameter to control the residual weight.

Specifically, for hidden layers not directly used as the output for downstream tasks, we use vector concatenation operation, as shown in Equation~\ref{eq:concat}. Otherwise, we use vector average operation to make the dimension independent to the number of attention heads. Motivated by~\cite{l2norm}, we add L2 normalization for these averaged outputs for downstream tasks. Formalization is shown in Equation~\ref{eq:l2} and~\ref{eq:l2_2}.

\begin{equation}
\label{eq:concat}
    \mathbf{h}^{(l)}_i = \concat_{k=1}^{T} \mathbf{o}^{k(l)}_i
\end{equation}

\begin{equation}
\label{eq:l2}
    \mathbf{h}'^{(l)}_i  = \frac{1}{T}
    \sum_{k=1}^{T} \mathbf{o}^{k(l)}_i
\end{equation}

\begin{equation}
\label{eq:l2_2}
    \mathbf{h}^{(l)}_i = \frac{\mathbf{h}'^{(l)}_i}{\lVert \mathbf{h}'^{(l)}_i \rVert}
\end{equation}

}

\section{Experiments}
\label{sec:exp}


We benchmark results for 1) all HGNNs discussed in Section \ref{sec:relwork}, 2) GCN and GAT, and 3) \smodel on HGB. 
All experiments are reported with the average and the standard variance of five runs. 

\subsection{Benchmark}

Tables~\ref{tab:NC}, \ref{tab:LP}, and \ref{tab:recom_benchmark} 
report results for node classification, link prediction, and knowledge-aware recommendation, respectively. 
The results show that under fair comparison, 
1) the simple homogeneous GAT can matches the best HGNNs in most cases, and 
2) inherited from GAT, \smodel consistently outperforms all advanced HGNNs methods for node classification on four datasets, link prediction on three datasets, and knowledge-aware recommendation on three datasets. 

Implementations of all previous HGNNs are \textbf{based on their official codes} to avoid errors introduced by re-implementation. 
The only modification occurs on their data loading interfaces and downstream decoders, if necessary, to make their codes adapt to the HGB pipeline.


We use Adam optimizer with weight decay for all methods, and tune hyperparameters based on the validation set performance. The details of hyperparameters are recorded in Appendix~\ref{app:hyper}. For the methods requiring meta-paths, the meta-paths used in benchmark datasets are shown in Appendix~\ref{app:metapath}.

\subsection{Time and Memory Consumption}

We test the time and memory consumption of all available HGNNs for node classification on the DBLP dataset. The results are showed in Figure~\ref{fig:tam}. It is worth noting that we only measure the time consumption of one epoch for each model, but the needed number of epochs until convergence could be various and hard to exactly define. HetSANN is omitted due to our failure to get a reasonable Micro-F1 score.

\subsection{Ablation Studies}
The ablation studies on all the three tasks are summarized in Table~\ref{tab:ablation}. Residual connection and $L_2$ normalization consistently improve performance, but type embedding only slightly boosts the performance on node classification, although it is the best way in our experiments to encode type information explicitly under the GAT framework. We will discuss the possible reasons in \S~\ref{sec:discuss}.

\hide{

\section{Experiments}
\label{sec:exp}

\hide{
\begin{table*}[]
\centering
\scriptsize
\begin{tabular}{c|c|c|c|c|c|c|c|c|c|c|c|c|c|c}

\toprule
        & \multicolumn{2}{c|}{HAN}  & \multicolumn{3}{c|}{GTN}    & \multicolumn{3}{c|}{RSHN} & \multicolumn{4}{c|}{HetGNN}                                 & \multicolumn{2}{c}{MAGNN}   \\ \hline
Dataset & \multicolumn{2}{c|}{ACM}  & DBLP  & ACM   & IMDB       & AIFB    & MUTAG  & BGS    & \multicolumn{2}{c|}{MC (10\%)}   & \multicolumn{2}{c|}{MC (30\%)}  & \multicolumn{2}{c}{DBLP}  \\ \hline
Metric  & Macro-F1   & Micro-F1     &     &         &             &         &        &        & Macro-F1    & Micro-F1         & Macro-F1   & Micro-F1           & Macro-F1      & Micro-F1 \\ 
\midrule
model*  & 91.89$\pm$0.38    & 91.85$\pm$0.38   & 94.18       & 92.68  & 60.92        & 97.22   & \textbf{82.35}  & 93.10  & 0.978     & 0.979   & 0.981                           & 0.982                           & 93.13                                    & 93.61                                    \\ 
GCN*    & 89.31$\pm$0.50                           & 89.45$\pm$0.45                           & 87.30                                    & 91.60                                    & 56.89                                    & -       & -      & -      & -                               & -                               & -                               & -                               & 88.00                                    & 88.51                                    \\ 
GAT*    & 90.55$\pm$0.59                           & 90.55$\pm$0.59                           & 93.71                                    & 92.33                                    & 58.14                                    & 91.67   & 72.06  & 66.32  & 0.962                           & 0.963                           & 0.965                           & 0.965                           & 91.05                                    & 91.61                                    \\ 
\midrule
model   & -                                        & -                                        & 92.95$\pm$0.37                           & 92.28$\pm$0.73                           & 57.53$\pm$2.22                           & -       & -      & -      & -                               & -                               & -                               & -                               & 92.81$\pm$0.43                           & 93.36$\pm$0.40                           \\ 
GCN     & 92.08$\pm$0.62                           & 92.15$\pm$0.55                           & 91.48$\pm$0.35                           & 92.28$\pm$0.16                           & \textbf{59.11$\pm$1.73} & 97.22   & 79.41  & 96.55  & 0.965                           & 0.964                           & 0.977                           & 0.978                           & 88.31$\pm$0.38                           & 89.37$\pm$0.29                           \\ 
GAT     & \textbf{92.47$\pm$0.24} & \textbf{92.45$\pm$0.25} & \textbf{94.18$\pm$0.14} & \textbf{92.49$\pm$0.65} & 58.86$\pm$1.73                           & \textbf{100}     & 80.88  & \textbf{100}    & \textbf{0.987} & \textbf{0.987} & \textbf{0.990} & \textbf{0.990} & \textbf{94.40$\pm$0.30} & \textbf{94.78$\pm$0.28} \\ \bottomrule
\end{tabular}
\end{table*}
}

\yx{any reason why GAT/GCN is not included int able 5 and 6}

\subsection{Benchmark}

We make a comprehensive comparison for current HGNNs on HGB. The results for node classification, link prediction and knowledge-aware recommendation are demonstrated in Table~\ref{tab:NC}, Table~\ref{tab:LP} and Table~\ref{tab:recom_benchmark} respectively. The results show that well-implemented GAT already matches the best HGNNs in most tasks, and \smodel outperforms other methods by a large margin.

Implementations of all previous HGNNs are \textbf{based on their official codes} to avoid errors introduced by re-implementation. We only modify their data loading interfaces and downstream decoders, if necessary, to make their codes adapt to the HGB pipeline.


We use Adam optimizer with weight decay for all methods, and tune hyperparameters based on the validation set performance. The details of hyperparameters are recorded in Appendix~\ref{app:hyper}. For the methods requiring meta-paths, the meta-paths used in benchmark datasets are shown in Appendix~\ref{app:metapath}.

\subsection{Time and Memory Consumption}
We test the time and memory consumption of all available HGNNs for node classification on the DBLP dataset. The results are showed in Figure~\ref{fig:tam}. It worth noting that we only measure the time consumption of one epoch for each model, but the needed number of epochs until convergence could be various and hard to exactly define. HetSANN is omitted due to our failure to get a reasonable Micro-F1 score.

\begin{figure}[h]
\includegraphics[width=\columnwidth]{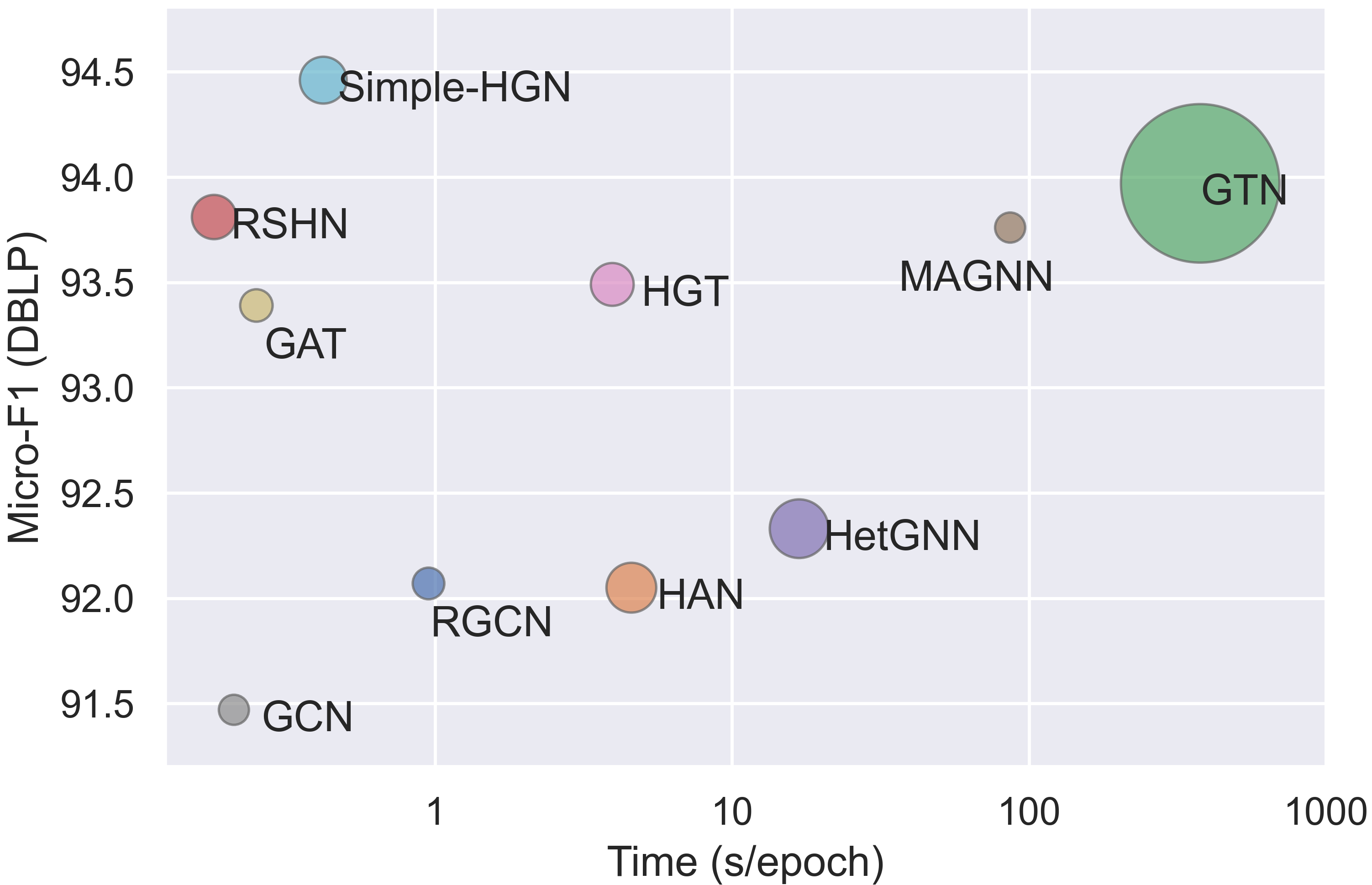}
\caption{Time and memory comparison for HGNNs on DBLP. The area of the circles represent the memory consumption of the corresponding models.}
\label{fig:tam}
\end{figure}

\subsection{Ablation Studies}
The ablation studies on all the three tasks are summarized in Table~\ref{tab:ablation}. Residual connection and $L_2$ normalization consistently improve performance, but type embedding only slightly boosts the performance on node classification, although it is the best way in our experiments to encode type information explicitly under the GAT framework. We will discuss the possible reasons in \S~\ref{sec:discuss}.

}
\section{Discussion and Conclusion}\label{sec:discuss}

In this work, we identify the neglected issues in heterogeneous GNNs, setup the heterogeneous graph benchmark (HGB), and introduce a simple and strong baseline \model. 
The goal of this work is to understand and advance the developments of heterogeneous GNNs by facilitating reproducible and robust research. 

Notwithstanding the extensive and promising results, there are still open questions remaining for heterogeneous GNNs and broadly heterogeneous graph representation learning. 

\vpara{Is explicit type information useful?} Ablation studies in Table~\ref{tab:ablation} suggest the type embeddings only bring minor improvements. 
We hypothesize that the main reason is that the heterogeneity of node features already implies the different node and edge types. Another possibility is that the current graph attention mechanism~\cite{velivckovic2017graph} is too weak to fuse the type information with feature information. We leave this question for future study.

\vpara{Are meta-paths or variants still useful in GNNs?} Meta-paths~\cite{Sun:BOOK2012} are proposed to separate different semantics with human prior. 
However, the premise of (graph) neural networks is to avoid the feature engineering process by extracting implicit and useful features  underlying the data. 
Results in previous sections also suggest that meta-path based GNNs do not generate outperformance over the homogeneous GAT. 
Are there better ways to leverage meta-paths in heterogeneous GNNs than existing attempts? 
Will meta-paths still be necessary for heterogeneous GNNs in the future and what are the substitutions?


\hide{
\section{Discussion}\label{sec:discuss}
Although the extensive experiments in this work reveal the issues and setup new benchmarks and baselines for HGNNs, some core problems as follows about heterogeneous graphs remain open.

\vpara{Is explicit type information useful?} Ablation studies in Table~\ref{tab:ablation} suggest the type embeddings only bring minor improvements. We hypothesize that the main reason is that the heterogeneity of node features already implies the different node and edge types. Another possibility is that the current graph attention mechanism~\cite{velivckovic2017graph} is too weak to fuse the type information with feature information. We leave this question for future study.

\vpara{Are meta-paths still useful in GNNs?} Meta-paths~\cite{sun2011pathsim} are proposed to separate different semantics with human prior. However, current GNNs have greater modeling power than
models 10 years ago, and meta-path based methods do not show better performance on HGB. Will meta-paths still be necessary for heterogeneous graphs in the future and what is the substitution?

\section{Conclusion}\label{sec:conclusion}
In this work, we reveals the neglected problems in current HGNNs, and setup new benchmark HGB and a strong baseline \model. HGB contains 11 datasets from 3 representative tasks, and will become a new arena for effective HGNN models. \smodel surpasses previous methods with three simple modification over GAT. The results of extensive experiments also drive us to question the necessity of customary techniques, e.g., meta-paths, which indicates the urgency to rethink heterogeneous graph modeling in the era of deep learning.

}

\begin{acks}
The work is supported by the NSFC for Distinguished Young Scholar (61825602) and NSFC (61836013). 
The authors would like to thank Haonan Wang from UIUC and Hongxia Yang from Alibaba for their kind feedbacks.
\end{acks}

\bibliographystyle{ACM-Reference-Format}
\bibliography{ref}

\clearpage
\appendix

\section{Time and Memory Consumption}

\begin{figure}[h]
\includegraphics[width=0.65\columnwidth]{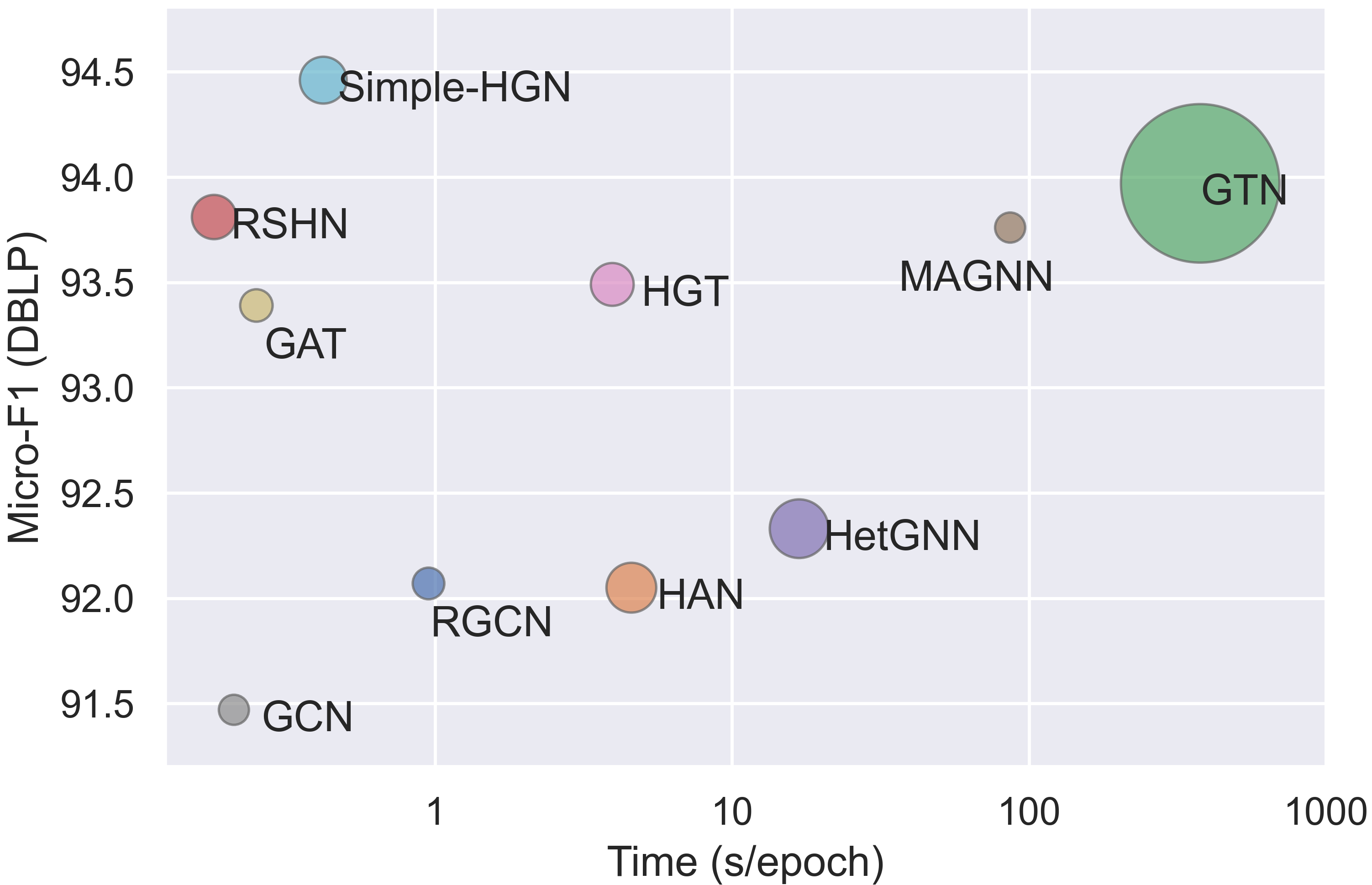}
\caption{Time and memory comparison for HGNNs on DBLP. The area of the circles represent the (relative) memory consumption of the corresponding models. }
\label{fig:tam}
\vspace{-0.3cm}

\end{figure}

\section{Random negative}\label{app:randneg}
The distribution of negative samples of test set in link prediction task has a great impact on the performance score. The results with random negative test in our benchmark are shown in Table~\ref{tab:random_neg}. As we can see, the scores are greater than those in Table~\ref{tab:LP} by a large margin. Most works~\cite{cen2019representation,magnn,hetgnn} evaluate link prediction with randomly sampled negative pairs, which are easy to distinguish from the positive pairs for most methods. However, in real world scenarios, we usually need to discriminate positive and negative node pairs with similar characters, instead of random ones, due to the widely used ``retrieve then re-rank'' industrial pipeline. Therefore, we choose to use sampled 2-hop neighbors as our negative test set in benchmark.

\begin{table*}
\caption{Link prediction benchmark with random negative test.}
\footnotesize
\label{tab:random_neg}
\begin{tabular}{ccccccc}
\toprule
           & \multicolumn{2}{c}{Amazon}                   & \multicolumn{2}{c}{Last.fm}                                     & \multicolumn{2}{c}{PubMed}                                      \\ \midrule
           & ROC-AUC               & MRR                  & ROC-AUC                        & MRR                            & ROC-AUC                        & MRR                            \\ \midrule
RGCN	&	89.76$\pm$0.33	&	95.76$\pm$0.22	&	81.90$\pm$0.29	&	96.68$\pm$0.14	&	88.32$\pm$0.08	&	96.89$\pm$0.20\\
GATNE	&	96.67$\pm$0.08	&	98.68$\pm$0.06	&	87.42$\pm$0.22	&	96.35$\pm$0.24	&	78.36$\pm$0.92	&	90.64$\pm$0.49\\
HetGNN	&	95.51$\pm$0.39	&	97.91$\pm$0.08	&	87.35$\pm$0.02	&	96.15$\pm$0.18	&	84.14$\pm$0.01	&	91.00$\pm$0.03\\
MAGNN	&	-	&	-	&	76.50$\pm$0.21	&	85.68$\pm$0.04	&	-	&	-\\
HGT	&	91.70$\pm$2.31	&	96.07$\pm$0.68	&	80.49$\pm$0.78	&	95.48$\pm$0.38	&	90.29$\pm$0.68	&	97.31$\pm$0.09\\ \midrule
GCN	&	98.57$\pm$0.21	&	\textbf{99.77$\pm$0.02}	&	84.71$\pm$0.1	&	96.60$\pm$0.12	&	86.06$\pm$1.23	&	98.80$\pm$0.56\\
GAT	&	98.45$\pm$0.11	&	99.61$\pm$0.22	&	83.55$\pm$2.11	&	91.45$\pm$5.66	&	87.57$\pm$1.23	&	98.38$\pm$0.11\\ \midrule
Simple-HGN & \textbf{98.74$\pm$0.25} & 99.52$\pm$0.08         & \textbf{91.04$\pm$0.22} & \textbf{99.21$\pm$0.15} & \textbf{91.40$\pm$0.30} & 96.04$\pm$0.25         \\         \bottomrule
\end{tabular}
\end{table*}

\section{Hyper-parameters}
\label{app:hyper}

We search learning rate within $\{1,5\}\times\{10^{-6},10^{-5}, 10^{-4}, 10^{-3}, 10^{-2}\}$ in all cases, and $\{0,1,2,5\}\times\{10^{-6},10^{-5}, 10^{-4}, 10^{-3}\}$ for weight decay rate. We set dropout rate as 0.1 in recommendation, and 0.5 in node classification and link prediction by default. For batch size, we will try $\{1024,2048,4096,8192\}$, unless the code of the author has special requirements. For training epoch, we will use early stop mechanism based on the evaluation on validation set to promise fully training.

For brevity, we will denote some variables. Suppose dimension of embeddings for graph layers as $d$, dimension of edge embeddings as $d_e$, dimension of attention vector (if exists) as $d_a$, number of graph layers as $L$, number of attention heads as $n_h$, negative slope of LeakyReLU as $s$.

For input feature type, we use $\text{feat}=0$ to denote using all given features, $\text{feat}=1$ to denote using only target node features, and $\text{feat}=2$ to denote all nodes with one-hot features.

\subsection{\model}

\subsubsection{Node classification}

We set $d=d_e=64$, $n_h=8$, $\beta=0.05$ for all datasets. For DBLP, ACM and Freebase datasets, we set $L=3$, $s=0.05$. For IMDB dataset, we set $L=6$, $s=0.1$. We set $\text{feat}=0$ for IMDB, $\text{feat}=1$ for ACM, and $\text{feat}=2$ for DBLP and Freebase.

\subsubsection{Link prediction}

We set $d=64$, $d_e=32$, $n_h=2$, $\beta=0$, $s=0.01$ for all datasets. For Amazon and PubMed, we use DistMult as decoder, and set $L=3$. For LastFM, we use dot product as decoder, and set $L=4$. We use $\text{feat}=2$ for all datasets.

\subsubsection{Recommendation}

For all datasets, we set $d^{(0)}=64$, $d^{(1)}=32$, $d^{(2)}=16$, $n_h=1$, $s=0.01$ as suggested in~\cite{wang2019kgat}.

\subsection{HAN}

\subsubsection{Node classification}
We set $d=8$, $d_a=128$, $n_h=8$ and $L=2$ for all datasets. For input feature type, we use $\text{feat}=2$ in Freebase, and $\text{feat}=1$ in other datasets. We have also tried larger $d$, but the vairation of performance becoms very large. Therefore, we keep $d=8$ as suggested in HAN's code.

\subsection{GTN}

\subsubsection{Node classification}
We use adaptive learning rate suggested in their paper for all datasets. We set $d=64$, number of GTN channels as 2. For DBLP and ACM, we set $L=2$. For IMDB dataset, we set $L=3$.

Moreover, as suggested in GTN paper, we aggregate the keyword node information as attribute to neighbors and use the left sub-graph to do node classification. We also tried to use the whole graph for GTN. Unfortunately, it collapse in that case, which indicates GTN is sensitive to the graph structure.

\subsection{RSHN}

For IMDB and DBLP, we set $L=2$ and $d=16$. For ACM, we $L=3$ and $d=32$. We use $\text{feat}=0,1,2$ for ACM, IMDB and DBLP respectively.

\subsection{HetGNN}

\subsubsection{Node classification}
We set $d=128$, $\text{feat}=0$, and batch size as 200 for all datasets. For random walk, we set walk length as 30 and the window size as 5.

\subsubsection{Link prediction}
We set $d=128$, $\text{feat}=2$, and batch size as 200 for all datasets. For random walk, we set walk length as 30 and the window size as 5.

\subsection{MAGNN}

We set $d=64$, $d_a=128$ and $n_h=8$ in all cases.

\subsubsection{Node classification}

We use $\text{feat}=1$ in all cases. For DBLP and ACM datasets, we set batch size as 8, and number of neighbor samples as 100. For IMDB dataset, we use full batch training.

\subsubsection{Link prediction}

We set batch size as 8, and number of neighbor samples as 100 for LastFM. For other datasets, we failed to adapt the MAGNN code to them because there is too much hard-coding.

\subsection{HetSANN}

\subsubsection{Node classification}

For ACM, we set $d=64$, $L=3$, $n_h=8$ and $\text{feat}=0$. For IMDB, we set $d=32$, $L=2$, $n_h=4$ and $\text{feat}=1$. For DBLP, we set $d=64$, $L=2$, $n_h=4$ and $\text{feat}=2$.

\subsection{HGT}
\subsubsection{Node Classification}
We use layer normalization in each layer, and set $d=64$, $n_h=8$ for all datasets.
$L$ is set to 2, 3, 3, 5 for ACM, DBLP, Freebase and IMDB respectively. For input feature type, we use $\text{feat}=2$ in Freebase, $\text{feat}=1$ in IMDB and DBLP, and $\text{feat}=0$ in ACM.
\subsubsection{Link Prediction}
For all datasets, we use layer normalization in each layer, and set $d=64$, $n_h=8$, $\text{feat}=2$ and DistMult as decoder. 

\subsection{GCN}

\subsubsection{Node classification}

We set $d=64$ for all datasets. We set $L=3$ for DBLP, ACM and Freebase, and $L=4$ for IMDB. We use $\text{feat}=2$ for DBLP and Freebase, and $\text{feat}=0$ for ACM and IMDB.

\subsubsection{Link prediction}

We set $d=64$, $L=2$, and $\text{feat}=2$ for all datasets.

\subsection{GAT}

\subsubsection{Node classification}

We set $d=64$, $n_h=8$ for all datasets. For DBLP, ACM and Freebase, we set $s=0.05$ and $L=3$. For IMDB, we set $s=0.1$ and $L=5$. We use $\text{feat}=2$ for DBLP and Freebase, $\text{feat}=1$ for ACM, and $\text{feat}=0$ for IMDB.

\subsubsection{Link prediction}

We set $d=64$, $n_h=4$, $L=3$ and $\text{feat}=2$ for all datasets.

\subsection{RGCN}
\subsubsection{Node classification}
We set $L=5$ for all datasets. For ACM, we set $d=16$, $\text{feats}=2$.
For DBLP and Freebase, we set $d=16$, $\text{feats}=3$. For IMDB, we set $d=32$, $\text{feats}=1$.

\subsubsection{Link prediction}
We set $L=4$, $d=64$ and $\text{feat}=2$ for all datasets.

\subsection{GATNE}

\subsubsection{Link prediction}

We set $d=200$, $d_e=10$, $d_a=20$, $\text{feat}=2$ for all datasets.

For random walk, we set walk length as 30 and the window size as 5. For neighbor sampling, we set negative samples for optimization as 5, neighbor samples for aggregation as 10.

\subsection{KGCN and KGNN-LS}

\subsubsection{Recommendation}

For all datasets, we set $d^{(0)}=64$ and $d^{(1)}=48$. We also tried to stack more graph layers, but performance deteriorates when we do that, which is also found in ~\cite{kgcn,kgnnls} experiments. We use sum aggregator because it has best overall performance as reported in ~\cite{kgcn}.

\subsection{KGAT}

\subsubsection{Recommendation}

We set $d^{(0)}=64$, $d^{(1)}=32$, $d^{(2)}=16$ for all datasets. For attention mechanism, we keep the Bi-Interaction aggregator according to their official code.

\section{Meta-paths}\label{app:metapath}
The meta-paths used in benchmark experiments are shown in Table~\ref{tab:metapath}. We try to select meta-paths following prior works. For example, meta-paths in DBLP, IMDB and LastFM are from~\cite{magnn}. Meta-paths in ACM dataset are based on~\cite{han}, and we also add some meta-paths related to citation and reference relation, which are not used in~\cite{han}. For Freebase dataset, we first count the most frequent meta-paths with length from 2 to 4, and then manually select 7 of them according to the performance on validation set.

\begin{table}
\caption{Meta-paths used in benchmark.}
\footnotesize
\label{tab:metapath}
\begin{tabular}{c|c|c}
\toprule
Dataset  & Meta-path                                                                   & Meaning                                                                                                                              \\ \midrule
DBLP     & \begin{tabular}[c]{@{}c@{}}APA\\APTPA\\APVPA\end{tabular}                    & \begin{tabular}[c]{@{}c@{}}A: author\\P: paper\\T: term\\V: venue\end{tabular}                                                       \\ \midrule
IMDB     & \begin{tabular}[c]{@{}c@{}}MDM, MAM\\DMD, DMAMD\\AMA, AMDMA\end{tabular}     & \begin{tabular}[c]{@{}c@{}}M: movie\\D: director\\A: actor\end{tabular}                                                              \\ \midrule
ACM      & \begin{tabular}[c]{@{}c@{}}PAP, PSP\\PcPAP, PcPSP\\PrPAP, PrPSP\end{tabular} & \begin{tabular}[c]{@{}c@{}}P: paper\\A: author\\S: subject\\c: citation relation\\r: reference relation\end{tabular}                     \\ \midrule
Freebase & \begin{tabular}[c]{@{}c@{}}BB\\BFB\\BLMB\\BPB\\BPSB\\BOFB\\BUB\end{tabular}  & \begin{tabular}[c]{@{}c@{}}B: book\\F: film\\L: location\\M: music\\P: person\\S: sport\\O: organization\\U: business\end{tabular}  \\ \midrule
LastFM   & \begin{tabular}[c]{@{}c@{}}UU, UAU\\UATAU\\AUA, ATA\\AUUA\end{tabular}       & \begin{tabular}[c]{@{}c@{}}U: user\\A: artist\\T: tag\end{tabular}   \\ \bottomrule                                                               
\end{tabular}
\end{table}

\end{document}